\documentclass{article}
\usepackage{ijcai22}

\usepackage{times}

\usepackage{soul}
\usepackage{url}
\usepackage[hidelinks]{hyperref}
\usepackage[utf8]{inputenc}
\usepackage[small]{caption}
\usepackage{graphicx}
\usepackage{amsmath}
\usepackage{booktabs}
\usepackage[ruled, vlined]{algorithm2e}
\SetKwComment{Comment}{/* }{ */}
\SetKwInput{kwInput}{Input}
\SetKwInput{kwOutput}{Output}
\SetKwInput{kwNote}{Comment}

\SetCommentSty{acommfont}

\usepackage{amsmath}
\usepackage{amssymb}
\usepackage{url}
\usepackage{booktabs}
\usepackage{multirow}
\usepackage{caption}
\usepackage{subcaption} 
\usepackage{pdflscape}
\usepackage[table]{xcolor}

\newtheorem{theorem}{Theorem}[section]

\newtheorem{example}[theorem]{Example}

\newcommand{\ov}[1]{\overline{#1}}

\newcommand{\SOTA}{SOTA}
\newcommand{\MSbasic}{MD}
\newcommand{\MSmAP}{AP}
\newcommand{\MSPred}{AP$\times$O}

\newcommand{\red}[1]{\textcolor{black}{{#1}}}

\title{ROAD-R: The Autonomous Driving Dataset with Logical Requirements}
\author{
Eleonora Giunchiglia$^1$\footnote{Contact authors.}\and
Mihaela C\u{a}t\u{a}lina Stoian$^{1*}$\and
Salman Khan$^2$\and \\
Fabio Cuzzolin$^2$\And
Thomas Lukasiewicz$^{3,1}$\\
\affiliations
$^1$Department of Computer Science, University of Oxford, UK\\
$^2$School of Engineering, Computing and Mathematics, Oxford Brookes University, UK\\
$^3$Institute of Logic and Computation, TU Wien, Austria\\
\emails 
eleonora.giunchiglia@cs.ox.ac.uk, mihaela.stoian@cs.ox.ac.uk, 19052999@brookes.ac.uk, fabio.cuzzolin@brookes.ac.uk, thomas.lukasiewicz@cs.ox.ac.uk
}
 
\begin{document}

\maketitle

\begin{abstract}
Neural networks have 
proven to be 
very powerful at
computer vision tasks.
However, 
they often exhibit unexpected behaviours, violating known requirements expressing background knowledge. 
This calls for models (i) able to learn from the requirements, and (ii) guaranteed to be compliant with the requirements  themselves. 
Unfortunately,
the development of such models is hampered by the lack of datasets equipped with formally specified requirements.
In this paper, we 
introduce the ROad event Awareness Dataset with logical Requirements (ROAD-R), the first 
publicly available dataset for autonomous driving with requirements expressed as logical constraints.
Given ROAD-R, we show that current state-of-the-art models 
often violate its logical constraints,
and that it is possible to exploit them to create models that (i) have a better performance, and (ii) are guaranteed to be compliant with the requirements themselves.

\end{abstract}

\section{Introduction}

Neural networks have 
proven to be incredibly powerful at processing low-level inputs, and for this reason they have been extensively applied to computer vision tasks, such as image classification, object detection, and action detection (see e.g., \cite{alexnet,yolo}).
However, they
can exhibit unexpected behaviors, contradicting known requirements expressing background knowledge. 
This can have dramatic consequences, 
especially in safety-critical scenarios such as autonomous driving.
To address the problem, 
models
should  (i) be able to learn from the requirements, and (ii) be guaranteed to be compliant with the requirements  themselves. 
Unfortunately, %
the development of such models is hampered by the lack of datasets equipped with formally specified requirements.
A notable exception is given by  hierarchical multi-label classification (HMC) problems (see, e.g., \cite{vens2008}) in which datasets are provided with binary constraints of the form $(A \to B)$ stating that label $B$ must be predicted whenever label $A$ is predicted.

In this paper, we introduce {\sl multi-label classification problems with propositional logic requirements}, in which datasets are provided with requirements ruling out non-admissible predictions and expressed in propositional logic. In this new formulation, given a multi-label classification problem with labels $A$, $B$ and $C$, we can, for example, write the requirement: 
$$
(\neg A \wedge B) \vee C,
$$
stating that for each datapoint in the dataset either the label $C$ is predicted, or $B$ but not $A$ are predicted. Obviously, any constraint written for HMC problems can be represented in our framework, and thus, our problem formulation represents a generalisation of HMC problems.

\red{Then, we present the ROad event Awareness Dataset with logical Requirements (ROAD-R), the first 
publicly available dataset for autonomous driving with requirements expressed as logical constraints.  ROAD-R extends the ROAD dataset \cite{gurkirt2021}, which consists
of 22 relatively long ($\sim8$ minutes each) videos annotated with {\sl road events}. A road event corresponds to
a tube, i.e., a sequence of frame-wise bounding boxes
linked in time. Each bounding box is labeled with a subset of the 41 labels specified in Table \ref{tab:labels}. 
The goal is to predict the set of labels associated to each bounding box.  
We manually annotated ROAD-R with 243 constraints, each 
verified to hold for each bounding box. A typical constraint is thus ``a traffic light cannot be red and green at the same time'', while there are no constraints like ``pedestrians should cross at
crossings”, which should always be satisfied in theory, but which might not be in real-world scenarios.
}

\red{
Given ROAD-R, we considered 6 current state-of-the-art (SOTA) models, and we showed that they are not able to learn the requirements  just from the data points, as more than 90\% of the times, they produce predictions that violate the constraints. Then, we
faced the problem of how to leverage the additional knowledge provided by constraints with the goal of (i)~improving their performance, measured by the frame mean average precision (f-mAP) at intersection over union (IoU) thresholds 0.5 and 0.75; see, e.g.,~\cite{kalogeiton2017a,li2018map}), and (ii) guaranteeing that they are compliant with the constraints. 
To achieve the above two goals, we propose the following new models: }
\begin{enumerate}
    \item 
    CL models, i.e., models with a {\sl constrained loss} allowing them to learn from the requirements,
    \item
    CO models, i.e, models with a {\sl constrained output} enforcing the requirements on the output, and
    \item
    CLCO models, i.e., models with both a constrained loss and a constrained output.
\end{enumerate}
\red{In particular, we consider three different ways to build CL (resp.,  CO, CLCO) models.
More specifically, we run the $9 \times 6$ models obtained by equipping the 6 current SOTA models with a constrained loss and/or a constrained output, and we show that it is always possible to}\red{
\begin{enumerate}
    \item improve the performance of each SOTA model, and 
    \item be compliant with (i.e., strictly satisfy) the constraints. 
\end{enumerate}}
Overall, the best performing model (for IoU = 0.5 and also IoU = 0.75) is CLCO-RCGRU, i.e., the SOTA model RCGRU equipped with both constrained loss and constrained output: CLCO-RCGRU (i)~always satisfies the requirements  and (ii)~has f-mAP = 31.81 for IoU = 0.5, and f-mAP = 17.27 for IoU = 0.75.
RCGRU, 
(i)~produces predictions that violate the constraints at least 92\% of the times,
and (ii) has f-mAP = 30.78 for IoU = 0.5 and f-mAP = 15.98 for IoU = 0.75. 

The main contributions of the paper thus are: 
\begin{itemize}
    \item we introduce multi-label classification problems with propositional logic requirements,
    \item we introduce ROAD-R,  which is the first publicly available dataset whose requirements are expressed in full propositional logic,
    \item we consider 6 SOTA models and show that on ROAD-R,  they produce predictions violating the requirements more than 90\% of the times,
    \item we propose new models with a constrained loss and/or constrained output, and show that in our new models, it is always possible to improve the performance of the SOTA models and satisfy the requirements.
\end{itemize}. 

\begin{table}[]
    \footnotesize
    \centering
    \begin{tabular}{l l l}
    \toprule
         \textbf{Agents} & \textbf{Actions} & \textbf{Locations} \\
    \midrule
    Pedestrian & Move away & AV lane \\
    Car & Move towards & Outgoing lane\\
    Cyclist & Move & Outgoing cycle lane \\
    Motorbike & Brake &  Incoming lane\\
    Medium vehicle & Stop & Incoming cycle lane \\ 
    Large vehicle & Indicating left & Pavement\\
    Bus & Indicating right & Left pavement \\
    Emergency vehicle & Hazards lights on & Right pavement\\
    AV traffic light & Turn left & Junction\\
    Other traffic light & Turn right & Crossing location \\
                        & Overtake & Bus stop \\
                        & Wait to cross & Parking\\
                        & Cross from left \\
                        & Cross from right \\
                        & Crossing \\
                        & Push object \\
                        & Red traffic light \\
                        & Amber traffic light \\
                        & Green traffic light \\
    \bottomrule
    \end{tabular}
    \caption{ROAD labels.}
    \label{tab:labels}
\end{table}

The rest of this paper is organized as follows. After the introduction to the problem, we present ROAD-R (Section \ref{sec:roadr}), followed by the evaluation of the SOTA models (Section \ref{sec:sota})  and of the SOTA models incorporating the requirements (Section \ref{sec:sotacorr}) on ROAD-R. We end the paper with the related work (Section \ref{sec:literature}) and the summary and outlook (Section \ref{sec:future}).

\section{%
Learning with Requirements
}\label{sec:prelim}

In ROAD, the detection of road events requires 
the following tasks: (i)~identify the bounding boxes,
(ii)~associate with each bounding box a set of labels, and (iii)~form a tube from the identified bounding boxes with the same labels.
Here, we focus on the second task, 
and we formulate it as a multilabel classification problem with requirements.

A {\sl multi-label classification} (MC) problem $\mathcal{P} \,{=}\, (\mathcal{C},\mathcal{X})$ consists of a finite set $\mathcal{C}$ of {\sl labels}, %
denoted by $A_1, A_2,\ldots$,
and a finite set $\mathcal{X}$ of pairs $(x,y)$, where $x \in \mathbb{R}^D$ $(D \ge 1)$ is a {\sl data point}, and $y \subseteq \mathcal{C}$ is the
{\sl ground truth} of~$x$. The ground truth $y$ associated with a data point $x$ characterizes both the  {\sl positive} and the {\sl negative labels}
 associated with $x$, defined to be $y$ and $\{\neg A: A \in \mathcal{C} \setminus y\}$, respectively. In ROAD-R, a data point corresponds to a bounding box, and each box is labeled with the positive labels representing (i)~the agent performing the actions in the box, (ii) the actions being performed, and (iii) the locations where the actions take place. See %
 Appendix~\ref{app:lab_description} for a detailed description of each label.
Consider an MC problem $\mathcal{P} = (\mathcal{C},\mathcal{X})$. A {\sl prediction} $p$ is a set of positive and negative labels such that for each label $A \in \mathcal{C}$, either $A \in p$ or $\neg A \in p$.
A {\sl model} $m$ for $\mathcal{P}$ is a function $m(\cdot,\cdot)$ mapping every label $A$ and every datapoint $x$ %
to $[0, 1]$. 
A datapoint $x$
is {\sl predicted} by $m$ to have label $A$ if its {\sl output value} $m(A,x)$ is greater than a user-defined {\sl threshold} $\theta \in [0, 1]$. The  {\sl prediction of $m$ for $x$} is the set 
$\{A : A \in \mathcal{C}, m(A,x) > \theta\} \cup \{\neg A : A \in \mathcal{C}, m(A,x) \leq \theta\}$ of positive and negative labels.

\begin{figure}[t]
    \centering
    \includegraphics[width=0.90\linewidth, trim={1.5cm 1.1cm 1.0cm 1.3cm},clip]{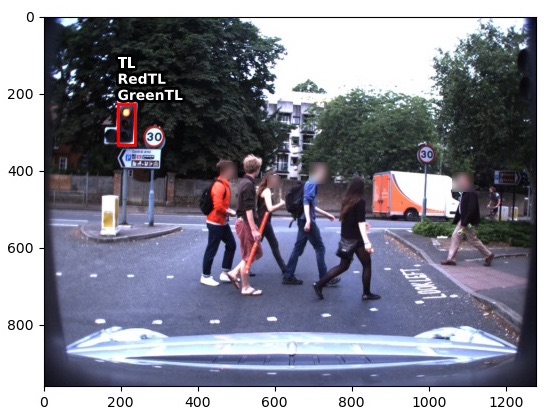}
    \caption{
    \!Example of violation of \,$\neg {\text{RedTL}}\,{\vee}\, \neg {\text{GreenTL}}$.}
    \label{fig:redtl}
\end{figure}

An {\sl MC problem with propositional logic requirements} $(\mathcal{P},\Pi)$ consists of an MC problem $\mathcal{P}$ and a finite set $\Pi$ of constraints ruling out non admissible predictions and expressed in propositional logic.

Consider an MC problem with requirements $(\mathcal{P},\Pi)$.
Each requirement delimits the set of  predictions that can be associated with each data point by ruling out those that  violate it. 
A prediction $p$ is 
{\sl admissible} if each constraint $r$ in $\Pi$ is {\sl satisfied} by $p$.
A model $m$ for $\mathcal{P}$ {\sl satisfies} (resp., {\sl violates}) the constraints on a data point $x$ if the prediction of $m$ for $x$ is (resp., is not) admissible.

\begin{example}\label{ex:1}\rm
The requirement that a traffic light cannot be both red and green  corresponds to the constraint 
 $\{\neg {\text{RedTL}},$ $\neg {\text{GreenTL}}\}$. Any prediction with
$\{\text{RedTL},$ $\text{GreenTL}\}$ 
is non-admissible. An example of such prediction is shown in Fig.~\ref{fig:redtl}.
\end{example}

Given an MC problem with requirements, it is possible to take advantage of the constraints in two different ways: %
\begin{itemize}
    \item they can be exploited during  learning  to teach the model the background knowledge that they express, and 
    \item they can be used as post-processing to turn a non-ad\-mis\-sible prediction into an admissible one. 
\end{itemize}
\red{Models in the first and second category are said to have a {\sl constrained loss} (CL) and {\sl constrained output} (CO) respectively.}
\red{Constrained loss models have the advantage
that the constraints are deployed during the training phase, and this should result in models (i) with a higher understanding of the problem and a better performance, but still (ii) with no guarantee that no violations will be committed}. 
On the other hand, \red{constrained output models
(i) do not exploit the additional knowledge during training, 
but (ii) are guaranteed to have no violations in the final outputs.} These two  options are not mutually exclusive (i.e., can be used together), and which one is to be deployed depends also on the extent to which a system is available. For instance, there can be companies that already have  their own models (which can be black boxes) and want to make them compliant with a set of requirements without modifying the model itself. On the other hand, the exploitation of the constraints in the learning phase can be an  attractive option for those who have a good knowledge of the model and want to further improve it.

\begin{table}[t]
    \centering
    \begin{tabular}{lrr}
    \toprule
    \textbf{Statistics} & & \\
    \midrule
      $|\mathcal{C}|$ & 41 \\
       $|\Pi|$ & 243 \\
       $\text{avg}_{r \in \Pi} (|r|)$ & 2.86 \\
       $|\{A \in \mathcal{C} : \exists r \in \Pi. A \in r \}|$ & 41 \\
       $|\{A \in \mathcal{C} : \exists r \in \Pi. \neg{A} \in r \}|$ & 38 \\
       $\text{min}_{A \in \mathcal{C}} (|\{r \in \Pi: \{A,\neg{A}\} \cap r \neq \emptyset\}|)$ & 2 \\        $\text{avg}_{A \in \mathcal{C}} (|\{r \in \Pi: \{A,\neg{A}\} \cap r \neq \emptyset\}|)$ & 16.95 \\ $\text{max}_{A \in \mathcal{C}} (|\{r \in \Pi: \{A,\neg{A}\} \cap r \neq \emptyset\}|)$ & 31 \\
    \bottomrule
    \end{tabular}
    \caption{Constraint statistics. All the constraints have between $2$ and~$15$ positive and negative labels, with an average of $2.86$. All (resp., $38$ of) the labels
    appear positively (resp., negatively) in $\Pi$. Each label 
    appears either positively or negatively between $2$ and $31$ times in $\Pi$, with an average of $16.95$.
    }
    \label{tab:req_stats}
\end{table}

\begin{figure*}[ht!]
\centering
    \begin{subfigure}[t]{0.32\textwidth}
    \includegraphics[width=0.98\textwidth,trim={0 0 7cm 7cm}]{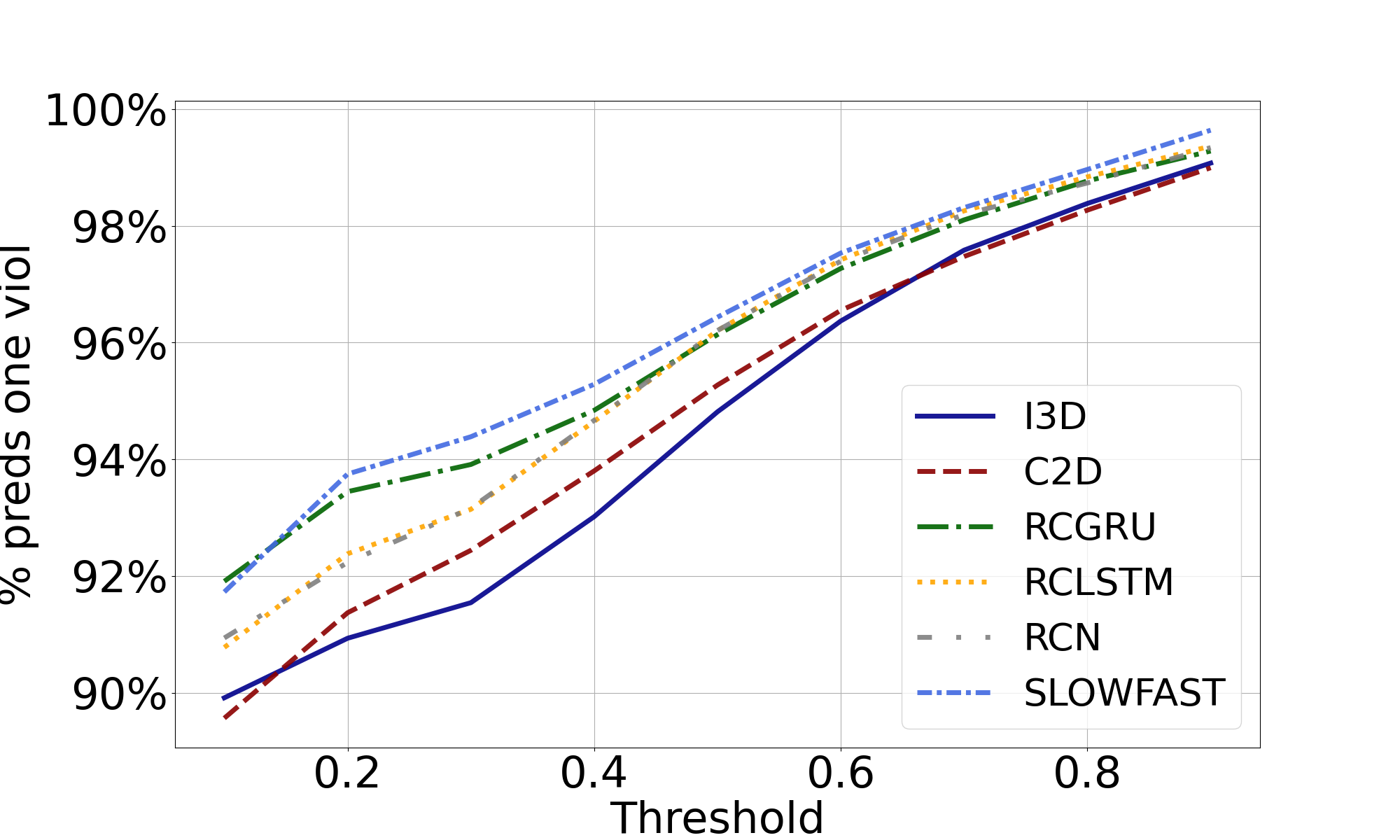}
    \caption{Percentage of predictions violating at least one constraint.} %
    \label{fig:sota-viola}
    \end{subfigure}
    \hfill
    \begin{subfigure}[t]{0.32\textwidth}
    \includegraphics[width=0.98\textwidth,trim={0 0 7cm 7cm}]{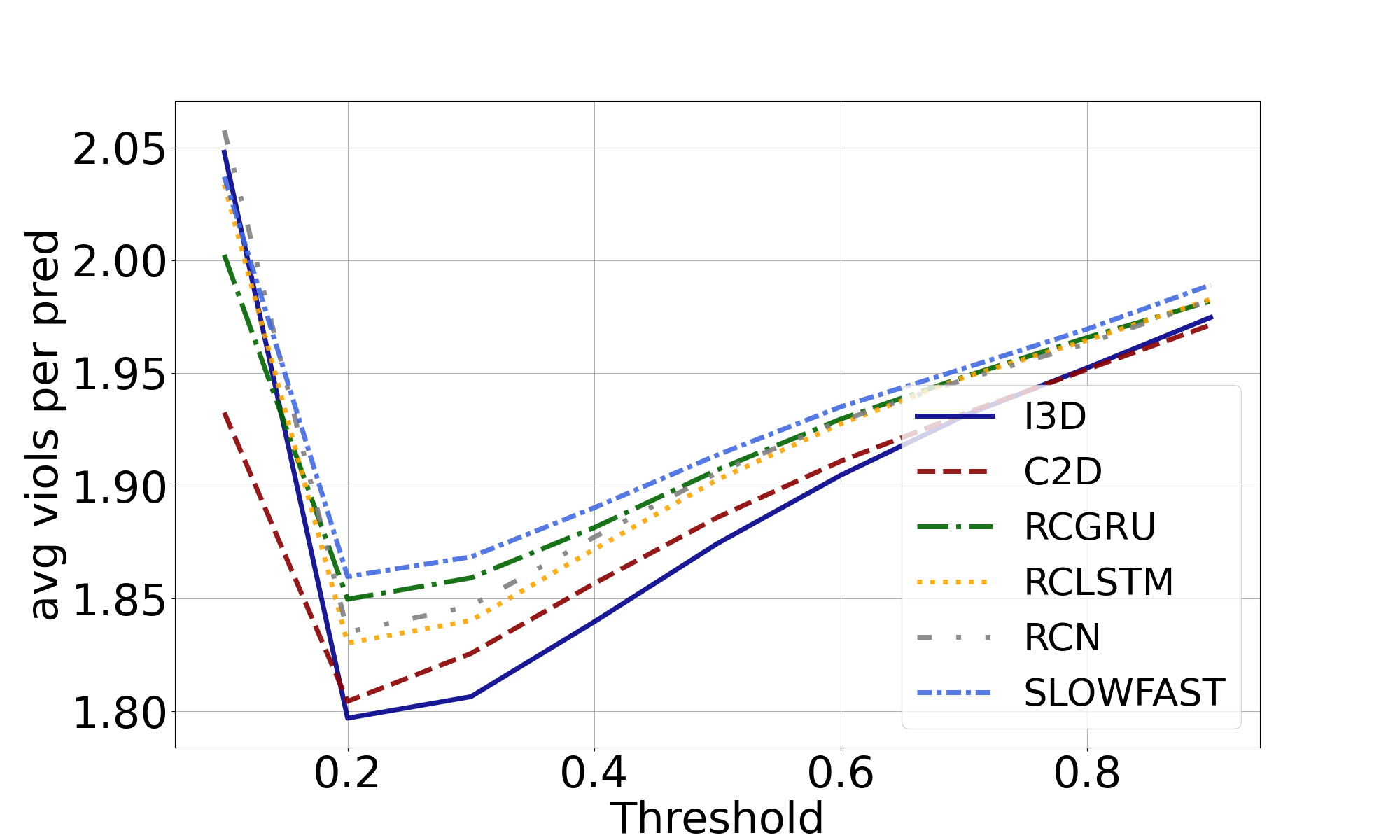}
    \caption{Average number of violations committed per prediction.} %
    \label{fig:sota-violb}
    \end{subfigure}
    \hfill
    \begin{subfigure}[t]{0.32\textwidth}
    \includegraphics[width=0.98\textwidth,trim={0 0 7cm 7cm}]{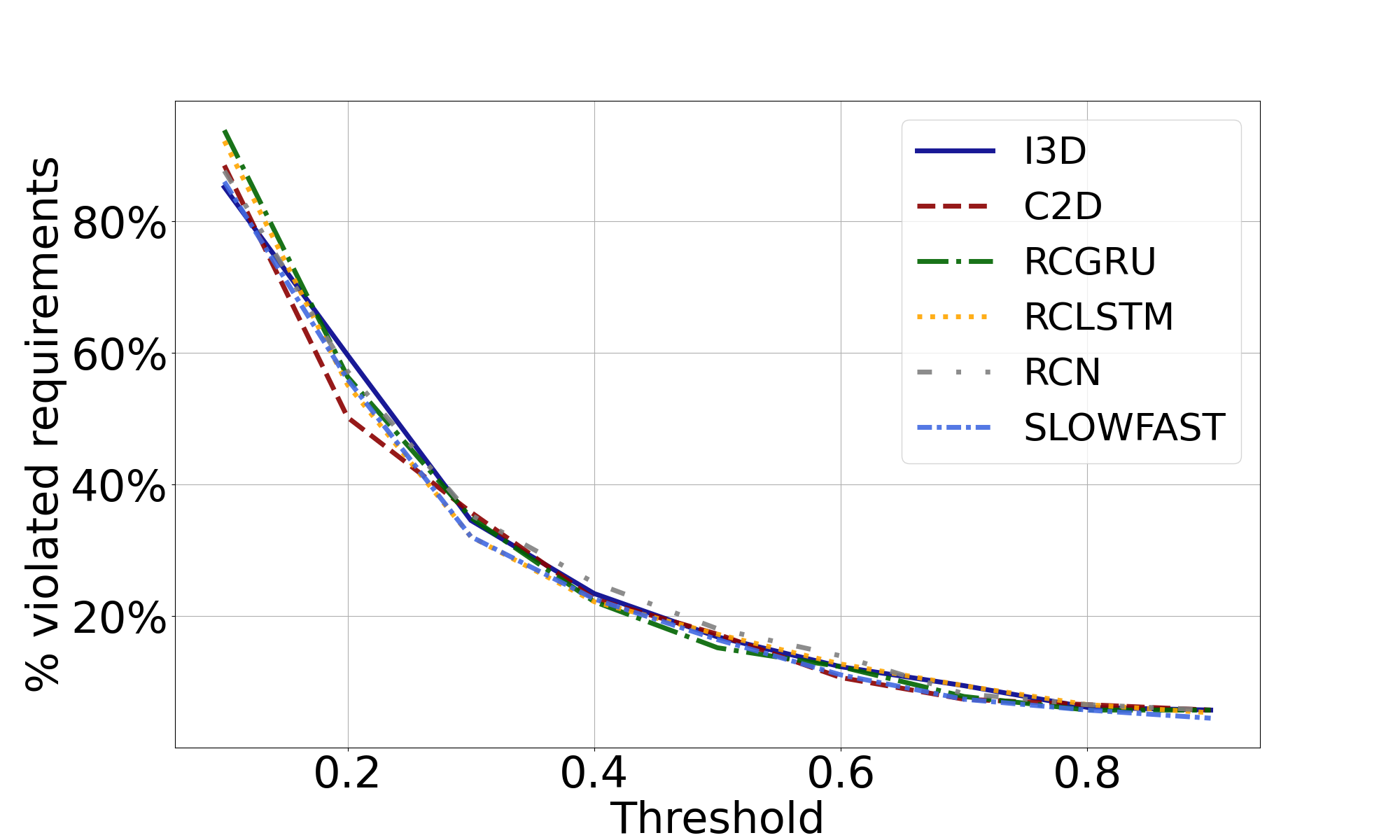}
    \caption{Percentage of constraints violated at least once.} %
    \label{fig:sota-violc}
    \end{subfigure}
    \caption{ROAD-R and SOTA models. In the $x$-axis, there is the threshold $\theta \in [0.1,0.9]$, step 0.1.}\label{fig:sota}
\end{figure*}

\section{ROAD-R}\label{sec:roadr}

ROAD-R extends the ROAD dataset\footnote{All the code will be released upon publication. 
ROAD is available at: https://github.com/gurkirt/road-dataset.}
\cite{gurkirt2021} by introducing a set  $\Pi$ of 243 constraints that specify the space of admissible outputs. In order to
improve the usability of our dataset, we write the constraints in a way that allows us to easily express $\Pi$ as a single formula in conjunctive normal form (CNF). The above can be done without any loss in generality, as any propositional formula can be expressed in CNF, and is important because many solvers expects formulas in CNF as input. Thus, each requirement in $\Pi$ has form: 
\begin{equation}\label{eq:req_lit}
l_1 \vee l_2 \vee \cdots \vee l_n\,, 
\end{equation}
where $n \geq 1$, and each $l_i$ is either a negative label $\neg A$ or a positive label~$A$. 
The requirements have been manually specified  following three steps: 
\begin{enumerate}
     \item 
an initial set of constraints $\Pi_1$ was manually created, 
     \item 
a subset $\Pi_2 \subset \Pi_1$ was retained by eliminating all those constraints that were entailed by the others,
     \item 
the final subset $\Pi \subset \Pi_2$ was retained by keeping only those requirements that were always satisfied by the ground-truth labels of the entire ROAD-R dataset.
\end{enumerate}

\begin{table}[t]
    \centering
    \begin{tabular}{rrrr}
    \toprule
$n$	&	$|\Pi_n|$ &	$\text{avg}_{r \in \Pi_n} (|r \cap \ov{\mathcal{C}}|)$ 	&	$\text{avg}_{r \in \Pi_n} (|r \cap \mathcal{C}|)$ \\
    \midrule
2	&	215	&	1.995	&	0.005 \\
3	&	5	&	1	&	2 \\
7	&	1	&	1	&	6\\
8	&	6	&	1	&	7\\
9	&	6	&	1	&	8\\
10	&	1	&	0	&	10\\
12	&	1	&	1	&	11\\
14	&	1	&	0	&	14\\
15	&	7	&	1	&	14\\
    \midrule
Total	&	243	&	1.87	&	0.96\\
    \bottomrule
    \end{tabular}
    \caption{Constraint statistics. $\Pi_n$ is the set of constraints $r$ in $\Pi$ with $|r| = n$, i.e., with $n$ positive and negative labels. $\ov{\mathcal{C}} = \{\neg {A} : A \in \mathcal{C}\}$. Each row shows the number of rules $r$ with $|r|=n$, and the average number of negative and positive labels in such rules.}
    \label{tab:req_stats1}
\end{table}

Finally, redundancy in the constraints has been automatically checked with {\sc relsat}\footnote{https://github.com/roberto-bayardo/relsat/}. 
Note that our process of gathering and further selecting the logical requirements follows more closely the software engineering paradigm rather than the machine learning view.
To this end, we ensured that the constraints were consistent with the provided labels from the ROAD dataset in the sense that they were acting as strict conditions to be absolutely satisfied by the ground-truth labels, as emphasized in the third step of the annotation pipeline above.
Tables~\ref{tab:req_stats} and \ref{tab:req_stats1} give a   high-level description of the properties of the set $\Pi$ of constraints. Notice that, with a slight abuse of notation, in the tables we use a set based notation for the requirements. Each requirement of form (\ref{eq:req_lit}) thus becomes 
$$
\{l_1, l_2, \ldots, l_n\}.
$$ Such notation allows us to express the properties of the requirements in a more succinct way.
In addition to the information in the tables, we report that of the 243 constraints, there are two in which all the labels are positive (expressing that there must be at least one agent and that every agent but traffic lights has at least one location), and 214 in which all the labels are negative (expressing mutual exclusion between two labels).  All the constraints with more than two labels have at most one negative label, as they express a one-to-many relation
 between actions and agents (like ``if something is crossing, then it is a pedestrian or a cyclist'').
Constraints like ``pedestrians should cross at crossings'', which might not be satisfied in practice, are not included.
Additionally embedding such logical constraints would require, e.g., using modal operators and, while it would be an interesting study to see the impact on the model's predictions when adding more complex layers to the expressivity of our logic, we opted for using a simpler logic in this first instance.
This also provides more transparency to the wider research community, as the full propositional logic covers a vast range of applications that do not require extra logical operators. 
The list with all the 243 requirements, with their natural language explanations, is  in Appendix~\ref{app:req_list},
Tables~\ref{tab:req_list1}, \ref{tab:req_list2}, and \ref{tab:req_list3}.
Notice that the 243 requirements restrict the number of admissible prediction to $4985868 \sim 5 \times 10^6$,
thus ruling out $(2^{41} - 4985868) \sim 10^{12}$ non-admissible predictions.\footnote{The number of admissible predictions has been computed with relsat: https://github.com/roberto-bayardo/relsat/.}
In principle, the set of admissible predictions can be further reduced by adding other constraints. \red{Indeed,} the 243 requirements are not guaranteed to be complete from every possible point of view: as standard in the software development cycle,
the requirement specification process deeply involves the stakeholders of the system (see, e.g., \cite{sommerville}). For example, we decided not to include constraints like ``it is not possible to both move towards and move away", which were not satisfied by all the data points because of errors in the ground truth labels. In these cases, we decided to dismiss the constraint in order to maintain (i) consistency between the knowledge provided by the constraints and by the data points, and (ii) backward compatibility. 

\red{As an additional point,} we underline that, even though the annotation of the requirements introduces some overhead in the annotation process, it is also the case that the effort of manually writing 243 constraints (i) is negligible when compared to the effort of manually annotating the 22 videos, and (ii) can improve such last process, e.g., allowing to prevent errors in the annotation of the data points.  %

\section{ROAD-R and SOTA Models}
\label{sec:sota}

As a first step, we ran 6 SOTA temporal feature learning architectures as part of a 3D-RetinaNet model~\cite{gurkirt2021} (with a 2D-ConvNet backbone made of Resnet50~\cite{he2016deep}) for event detection and evaluated 
to which extent constraints are violated. We considered:
\begin{enumerate}
\item {\sl 2D-ConvNet} (C2D)~\cite{wang2018non}:
     a Resnet50-based architecture with an additional temporal dimension for learning features from videos. The extension from 2D to 3D is done by adding a pooling layer over time to combine the spatial features.
\item {\sl Inflated 3D-ConvNet} (I3D)~\cite{carreira2017quo}:
      a sequential learning architecture extendable to any SOTA image classification model (2D-ConvNet based), 
    able to learn continuous spatio-temporal features from the sequence of frames.
\item {\sl Recurrent Convolutional Network} (RCN)~\cite{singh2019recurrent}:
     a 3D-ConvNet model that relies on recurrence for learning the spatio-temporal features at each network level. During the feature extraction phase, RCNs exploit both 2D convolutions across the spatial domain and
    1D convolutions across the temporal domain.  
\item {\sl Random Connectivity Long Short-Term Memory} (RCLSTM) \cite{hua2018traffic}:
     an updated version of LSTM in which the neurons are connected in a stochastic manner, rather than fully connected. In our case, the LSTM cell is used as a bottleneck in Resnet50 for learning the features sequentially.
\item {\sl Random Connectivity Gated Recurrent Unit} (RCGRU)  \cite{hua2018traffic}:
    an alternative version of RCLSTM where the GRU cell is used instead of the LSTM one. GRU makes the process more efficient with fewer parameters than the LSTM.
\item {\sl SlowFast} \cite{feichtenhofer2019slowfast}: a 3D-CNN architecture that contains both slow and fast pathways for extracting the sequential features. A Slow pathway computes the spatial semantics at low frame rate while a Fast pathway processes high frame rate for capturing the motion features. Both of the pathways are fused in a single architecture by lateral connections.
\end{enumerate}
We trained 3D-RetinaNet\footnote{https://github.com/gurkirt/3D-RetinaNets.} using the same hyperparameter settings for all the models: (i) batch size equal to 4, (ii)~sequence length equal to 8, and (iii) image input size equal to $512\times682$. All the models were initialized with the  Kinetics pre-trained weights. An SGD optimizer~\cite{lecun2012} with step learning rate was used. The initial learning rate was set to 0.0041 for all the models except SlowFast, for which it was set to 0.0021 due to the diverse nature of slow and fast pathways. All the models were trained for 30 epochs and the learning rate was made to drop by a factor of 10 after 18 and 25 epochs. The machine used for the experiments has 64 CPUs (2.2 GHz each) and 4 Titan RTX GPUs having 24 GB of RAM each. 

\begin{figure*}[t]

\begin{subfigure}[b]{0.33\textwidth}
         \centering
         \includegraphics[width=\textwidth,height=5.4cm, trim={50pt 30pt 50pt 10pt},clip]{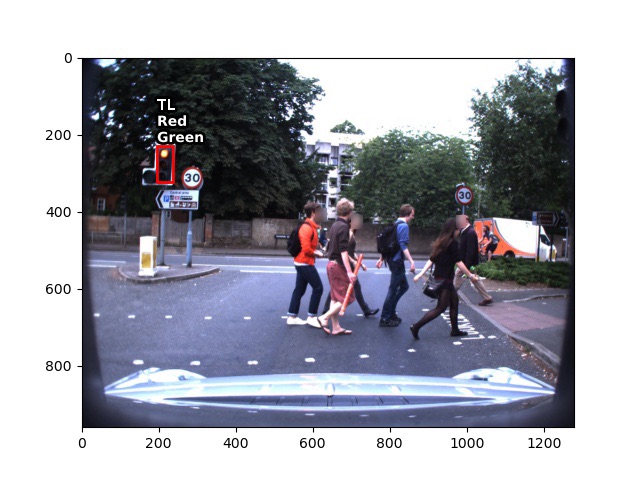}
         \caption{I3D}
         \label{I3D1}
\end{subfigure}
\begin{subfigure}[b]{0.33\textwidth}
         \centering
         \includegraphics[width=\textwidth, height=5.4cm, trim={20pt 20pt 50pt 10pt},clip]{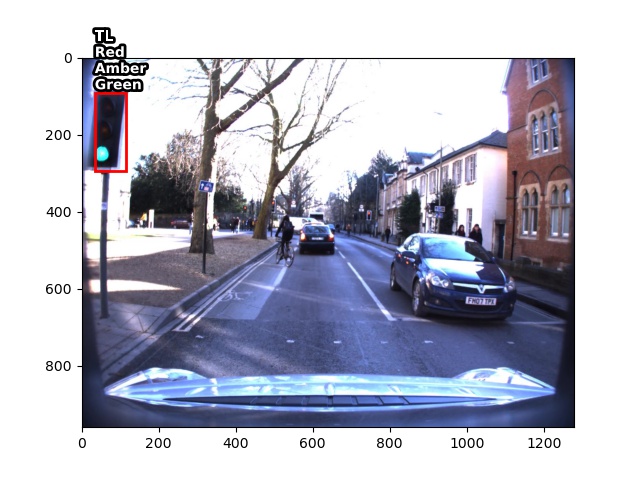}
         \caption{C2D}
         \label{C2D1}
\end{subfigure}
\begin{subfigure}[b]{0.33\textwidth}
         \centering
         \includegraphics[width=\textwidth, height=5.4cm, trim={20pt 20pt 50pt 10pt},clip]{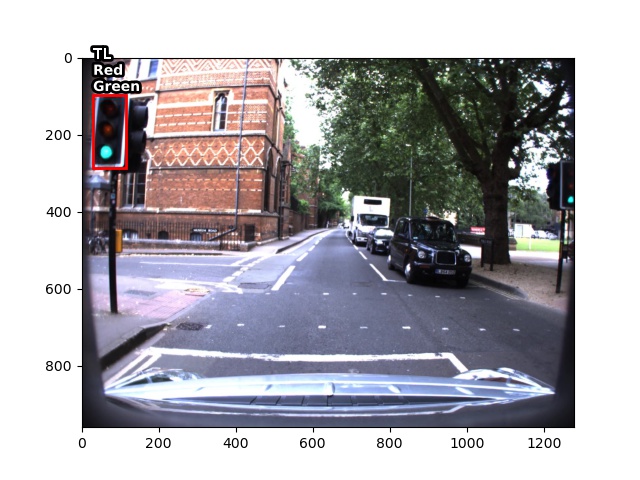}
         \caption{RCGRU}
         \label{RCGRU1}
\end{subfigure}

\begin{subfigure}[b]{0.33\textwidth}
         \centering
         \includegraphics[width=\textwidth, height=5.4cm, trim={40pt 20pt 40pt 10pt},clip]{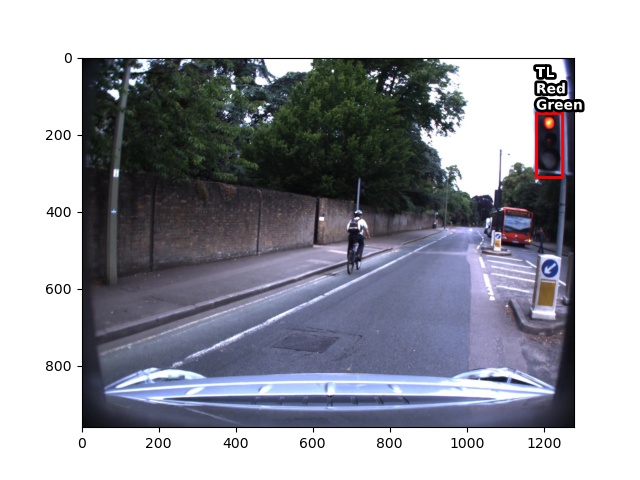}
         \caption{RCLSTM}
         \label{RCLSTM1}
\end{subfigure}
\begin{subfigure}[b]{0.33\textwidth}
         \centering
         \includegraphics[width=\textwidth, height=5.4cm, trim={20pt 20pt 50pt 10pt},clip]{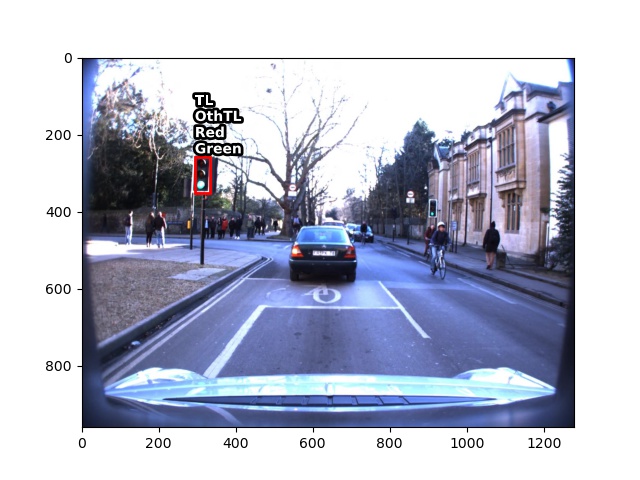}
         \caption{RCN}
         \label{RCN1}
\end{subfigure}
\begin{subfigure}[b]{0.33\textwidth}
         \centering
         \includegraphics[width=\textwidth,  height=5.4cm, trim={20pt 20pt 50pt 10pt},clip]{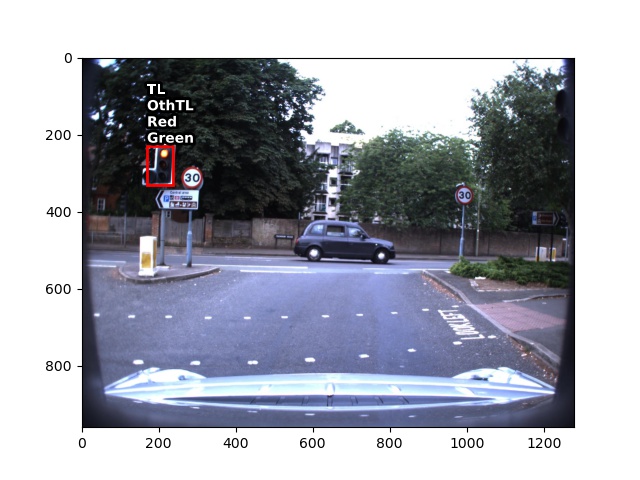}
         \caption{SlowFast}
         \label{Slowfast}
\end{subfigure}
\caption{Examples of violations of $\{\neg{\text{RedTL}},\neg{\text{GreenTL}}\}$.}
\label{fig:violRedTL}
\end{figure*}

To measure the models' performance, we used the {\sl frame mean average precision} (f-mAP), which is the standard metric used for action detection (see, e.g.,~\cite{kalogeiton2017a,li2018map}), with IoU threshold equal to 0.5 and 0.75, indicated as f-mAP@0.5 and f-mAP@0.75,  respectively. 
The results for the SOTA models at IoU threshold 0.5 and 0.75 are reported in Table \ref{tab:map5}, 
column ``SOTA''.

 To measure the extent to which each system violates the constraints, we used the following metrics: 
 \begin{itemize}
     \item the percentage of non-admissible predictions,
     \item the average number of violations committed per prediction, and
     \item  the percentage of constraints violated at least once,
 \end{itemize}
while varying the threshold $\theta$ from 0.1 to 0.9 with step 0.1. 
The results are in Fig.~\ref{fig:sota}, where (to improve readability) we do not plot the values corresponding to $\theta=0.0$ and $\theta=1.0$. For $\theta=0.0$ (resp., $\theta = 1.0$), all the predictions are positive (resp., negative), and thus the corresponding values are (in order) 100\%, 214, and 214/243 (resp., 100\%, 2, and 2/243).

Consider the results in Table \ref{tab:map5}, 
column ``\SOTA'', and in Fig.~\ref{fig:sota}. 
First, note that the performances are not an indicator of the ability of the model to \red{satisfy} the constraints. Indeed, higher f-mAPs do not correspond to lower trends in the plots of Fig.~\ref{fig:sota-violb}. For example,
RCGRU performs better than C2D for both IoU = 0.5 and IoU = 0.75, however, its curve is above C2D's in both Figs.~\ref{fig:sota-viola} and~\ref{fig:sota-violb}.
Then, note that the percentage of non-admissible predictions is always very high \red{for every model}: at its minimum, for $\theta=0.1$, more than 90\% of the predictions are non-admissible, and this percentage reaches 99\% for $\theta = 0.9$ (see Fig.~\ref{fig:sota-viola}). In addition, most predictions violate roughly two constraints, as shown  by Fig.~\ref{fig:sota-violb}. 
Considering that we are in an autonomous vehicle setting, such results are critical: one of the constraints that is  violated  
by all the baseline models
is $\{\neg{\text{RedTL}}, \neg{\text{GreenTL}}\}$, corresponding to predictions according to which there is a traffic light with both the red and the green lights on. Fig.~\ref{fig:violRedTL} shows an image for each of the SOTA models where such a prediction (for $\theta=0.5$) is made.  Appendix~\ref{app:imgs} contains qualitative examples of all the SOTA models making  predictions violating other constraints.

\begin{table*}[t]
\setlength{\tabcolsep}{2.4pt}
    \centering
    \footnotesize
    \begin{tabular}{l|c | c c c | c c c | c c c}
    \toprule
     & & \multicolumn{3}{c|}{{\textbf{Constrained Loss \red{(CL)}}}} &  \multicolumn{3}{c|}{{\textbf{Constrained Output \red{(CO)}}}} &  \multicolumn{3}{c}{{\textbf{Constrained Loss and Output \red{(CLCO)}}}} \\
         \textbf{Model}  & \textbf{\red{SOTA}} & \textbf{Product} & \textbf{G\"odel} & \textbf{Lukasiewicz}   &  \textbf{\MSbasic} & \textbf{\MSmAP} & \textbf{\MSPred} & \textbf{(P, AP$\times$O)} & \textbf{(G, AP$\times$O)} & \textbf{(L, AP$\times$O)} \\
         \midrule
         C2D & 27.57 & 27.90 \,\,(1)\, & 27.97 \,\,(1)\, & 27.57 (10)  & 27.41 (0.4) & 27.67 (0.9) & 27.69 (0.9) & 27.90 (0.7) & \textbf{28.16} (0.8) & 27.86 (0.9) \\
         I3D & 30.12 & 30.79 (10) & 29.98 (10) &  31.03 (10) & 29.86 (0.4) & 30.15 (0.5) & 30.15 (0.5) & 30.77 (0.9) & 30.02 (0.7) & \textbf{31.21} (0.6) \\
         RCGRU &  30.78 &  30.67 (10) & 29.57 \,\,(1)\, & {31.78} (10) & 30.56 (0.4) & {30.78} (0.5) & {30.78} (0.7) & 30.79 (0.6) & 29.83 (0.9) & \textbf{31.81} (0.5) \\
         RCLSTM & 30.49 & 30.48 (10) & 30.88 \,\,(1)\, & {31.63} (10) & 30.09 (0.4)&  {30.51} (0.7)  &  {30.51} (0.7) & 30.50 (0.5) & 30.90 (0.8) & \textbf{31.65} (0.9)  \\
         RCN & 29.64 & 30.31 (10) & 30.24 (10) & {\textbf{31.02}} (10) & 29.39 (0.4) & 29.80 (0.5)  & {29.82} (0.5) & 30.34 (0.6) & 30.28 (0.8) & \textbf{31.02} (0.5) \\
         SlowFast & 28.79 & 28.54 (10) & {28.91} \,\,(1)\, & 28.47 (10) & 28.51 (0.5) & {28.86} (0.9)  & {28.86} (0.9) & 28.67 (0.9) & \textbf{28.98} (0.5) & 28.56 (0.7) \\
         \midrule
         Avg. Rank & 7.08 & 5.75 &	5.50 &	4.25 & 	9.50 & 	5.75 & 	5.42 & 	4.42 &	4.50 & \textbf{2.83}  \\
    \bottomrule
    \multicolumn{3}{c}{}\\
    \end{tabular}
    
    \begin{tabular}{l|c | c c c | c c c | c c c}
    \toprule
     & & \multicolumn{3}{c|}{{\textbf{Constrained Loss \red{(CL)}}}} &  \multicolumn{3}{c|}{{\textbf{Constrained Output \red{(CO)}}}} &  \multicolumn{3}{c}{{\textbf{Constrained Loss and Output \red{(CLCO)}}}} \\
         \textbf{Model}  & \textbf{\red{SOTA}} & %
         \textbf{Product} & \textbf{G\"odel} & \textbf{Lukasiewicz}   &  \textbf{\MSbasic} & \textbf{\MSmAP} & \textbf{\MSPred} & \textbf{(P, AP$\times$O)} & \textbf{(G, AP$\times$O)} & \textbf{(L, AP$\times$O)} \\
         \midrule
  C2D & 12.66 & {13.04} \,\,(1)\, & 12.90 \,\,(1)\, & 12.30 (10) & 12.62 (0.4) & {12.72} (0.9) & {12.72} (0.9) & \textbf{13.05}  (0.8) & 13.01 (0.8) & 12.55 (0.9)  \\
         I3D & 16.29 & {16.47} (10) & 16.19 (10) & 16.32 (10) & 16.22 (0.5) & {16.31} (0.8) & {16.31} (0.8) & \textbf{16.48} (0.6) & 16.20 (0.6) & 16.42 (0.6) \\
         RCGRU &  15.98 & 16.57 (10) & 15.39 \,\,(1)\, & {17.26} (10) & 15.97 (0.4) & 16.02 (0.5) & {16.03} (0.5) & 16.63 (0.6) & 15.63 (0.9) & \textbf{17.27} (0.6) \\
         RCLSTM & {16.64}  & 16.14 (10) & 15.65 \,\,(1)\, & 16.34 (10) & 16.38 (0.4) &  \textbf{16.69} (0.5) & 16.66 (0.7) &  16.15 (0.7)  & 15.67 (0.8) & 16.39 (0.8) \\
         RCN &  15.82 & 16.57 (10) & 16.17 (10) & {17.25} (10) & 15.73 (0.4) & 15.87 (0.8) & {15.89} (0.5) & 16.56 (0.5) & 16.19 (0.9) & \textbf{17.26} (0.9) \\
         SlowFast &  14.40  & 14.84 (10) & 15.38 \,\,(1)\, & 13.80 (10) & 14.33 (0.5) & {14.47} (0.9) & {14.47} (0.9)  & {14.94} (0.9) & \textbf{15.42} (0.5) & 13.84 (0.7) \\
         \midrule
         Avg. Rank & 6.67	& 3.83 &	7.00 & 	5.67 & 	7.83 & 	5.00 &	4.83 & 	\textbf{3.17} & 	6.00 &	4.50  \\
         \bottomrule
\end{tabular}

    \caption{
f-mAP@0.5 (top table) and f-mAP@0.75 (bottom table) 
    for the 
    \red{
    (i) current SOTA models; (ii) CL models, in parentheses the value of $\alpha$; (iii) CO models, in parenthesis the threshold used to evaluate the admissibility of the predictions; and (iv) CLCO models}, in parenthesis the threshold used to evaluate the admissibility of the predictions. \textbf{P},  \textbf{G},  and \textbf{L} stand for the Product, G\"odel, and Lukasiewicz t-norm, respectively.
}
    \label{tab:map5}
\end{table*}

\section{ROAD-R and CL, CO, and CLCO Models}
\red{We now show how it is possible to build CL, CO and CLCO models. In particular, we show how to equip the 6 considered SOTA models with constrained a loss and/or a constrained output.}
\red{As anticipated in the introduction, we introduce (i) three different methods to build the constrained loss, (ii) three different methods to obtain the constrained output, and (iii) three combinations of constrained loss and constrained output.
Thus, we get 9 models for each SOTA model, for a total of 54. In order to get an overall view of the performance gains produced by each method, we also report the average ranking of the 9 proposed methods and SOTA
}
\cite{demsar2006}, computed as follows: \red{
\begin{enumerate}
    \item for each row in Table~\ref{tab:map5}, we rank the performances of the 9 CL, CO and CLCO models and of the SOTA model separately: the best performing \red{model} gets
the rank 1, the second best gets rank 2, etc., and in case of ties, the rank is split (e.g., the assigned rank is 1.5 if two \red{models} have the best performance), and 
\item for each column, %
we
take the average of the rankings computed in the previous step.
\end{enumerate}}
See Table \ref{tab:map5} 
for f-mAP@0.5, f-mAP@0.75 and average rankings.
In the table, for each row the best results  are in bold. 
The details of the implemented models with constrained loss, constrained output, and  both constrained loss and constrained output is given in the three paragraphs below.

\paragraph{Constrained Loss.} To constrain the loss, we take inspiration from the approaches proposed in~\cite{diligenti2017icmla,diligenti2017semantic}, and we %
train the models using the standard localization and classification losses, to which we add a regularization term. This last term represents the degree of satisfaction of the constraints in $\Pi$ and has the form: 
$$
\mathcal{L}_{\Pi} = \alpha \sum\nolimits_{i=1}^{|\Pi|} (1 - t(r_i)), 
$$
where $r_i$ represents the $i$th constraint in $\Pi$, $t(r_i)$ represents the fuzzy logic relaxation of $r_i$, and $\alpha$ is a hyperparameter ruling the weight of the regularization term (the higher $\alpha$ is, the more relevant the term corresponding to the constraints becomes, up to the limit case in which $\alpha \to \infty$ and the constraints become hard \cite{diligenti2017semantic}). 
We considered $\alpha \in \{1, 10, 100\}$ and the three fundamental t-norms:
\begin{enumerate}
    \item Product t-norm, 
    \item G\"odel t-norm, and
    \item Lukasiewicz t-norm
\end{enumerate}
as fuzzy logic relaxations (see e.g., \cite{metcalfe2005}).
The best results for f-mAP@0.5 and f-mAP@0.75 while varying $\alpha$ are in Table~\ref{tab:map5}, %
columns Product, G\"odel, and Lukasiewicz. As can be seen, \red{ SOTA  never achieves} the best average ranking, even when \red{compared with only the three CL methods. Of these,} Lukasiewicz (for IoU = 0.5) and Product (for IoU = 0.75) have the best ranking, though for some model and IoU, the best performances are obtained with G\"odel. In only one case (for RCLSTM at IoU = 0.75), the 
\red{SOTA} model performs better than the \red{CL} models. Also notice that the best performances are never obtained with $\alpha=100$, and that we never get any significant reduction in the number of predictions violating the constraints.

\paragraph{Constrained Output.} \label{sec:sotacorr}

We now consider the problem of how to correct a prediction $p$ whose admissibility is evaluated at a given threshold $\theta$. The first observation is that determining the existence of an admissible prediction is an intractable problem: indeed, this is just a reformulation of the satisfiability problem in propositional logic, which is well known to be NP-complete. Despite this, we want to correct any non-admissable prediction $p$ in such a way that (i) the final prediction is admissible, and (ii) the performance of the final model either improves or remains unaltered. 

In order to achieve the above, 
we first test the policy of trying to correct as few labels as possible. 
More precisely, for each prediction $q$, $(p \setminus q)$ is the set of positive and negative predictions on which $q$ differs from $p$. Then, we can compute the admissible prediction $q$ with the minimum number of differences, i.e., such that $| p \setminus q |$ is minimal. We call such policy {\sl Minimal Distance} (MD). Unfortunately, no polynomial time algorithm is known to solve this problem.
 \begin{theorem}\label{th:one}
 Let $(\mathcal{P},\Pi)$ be an MC problem with requirements. Let $p$ be a prediction. For each positive $d$, determining the existence of an admissible prediction $q$ such that $\mid p \setminus q \mid \le d$ is an NP-complete problem.
  \end{theorem}
The theorem is an easy consequence of Proposition 1 in \cite{DBLP:BailleuxM06}. 
In order to be able to solve the problem in practice, we formulate the problem of finding an admissible prediction with minimal $\mid p \setminus q \mid$ as a weighted partial maximum satisfiability (PMaxSAT) problem of a set of clauses\footnote{In our setting, a {\sl clause} is a disjunction of literals, and a {\sl literal} is either a positive label or its negation, representing the corresponding negative label. A clause is {\sl unit} if it consists of a single literal.} (see, e.g., \cite{DBLP:LiM09}) in which
\begin{enumerate}
\item each constraint in $\Pi$ corresponds to a clause marked as hard, and
\item each positive and negative prediction in $p$ corresponds to a unit clause marked as soft with unitary weight.
\end{enumerate}
This allows us to use the very efficient solvers publicly available for PMaxSat problems. In particular, in our experiments we used  MaxHS~\cite{hickey2019}, and running times were in the order of  $10^{-3}$s at most.
As intended, since we assign all labels unitary weight, MaxHS  
returns the admissible prediction $q$ with as few as possible labels flipped.
Notice that flipping the $i$th label amounts to changing its output value $o_i$ from a value below the threshold to another value $f(o_i)$ above the threshold or vice versa. In all our experiments, we considered (i)
$f(o_i) = \theta + \epsilon$, if $o_i < \theta$, and (ii) $f(o_i) = \theta - \epsilon$, otherwise ($\epsilon= 10^{-3}$). In this way, assuming $o_i \leq \theta$ (i.e., the label is negatively predicted and then flipped positive), we expect $f(o_i)$ to be lower (but still higher than $\theta$) than the output values of the non-flipped, positively predicted labels. It is done analogously for the case $o_i > \theta$.
We tested this approach with all the thresholds $\theta$ from 0.1 to 0.9 with step 0.1, 
and the resulting f-mAP@0.5 and f-mAP@0.75 for the best threshold are reported in Table~\ref{tab:map5},
column \red{MD}. As we can see from  the table, despite the fact that we are minimizing the number of corrections, the results obtained by the CO-MD model are always worse than the ones obtained by the SOTA models. We can hence conclude that adding such post-processing has a detrimental effect on the models' performance.

Thus, we need alternative policies to correct a non-admissable prediction $p$. We generalise the problem by assigning a positive weight $w_i$ (representing the cost of correcting the $i$th  label in $p$) and then computing the admissible prediction $q$ that minimizes $\sum_{i=1}^{41} w_i$. More precisely, for every prediction $q$, $cost(p,q) = \sum_{i \in \mathcal{I}} w_i$, $\mathcal{I}$ being the set of indexes of the labels in $(p \setminus q)$. Then, we can compute the admissable prediction $q$ such that $cost(p,q)$ is minimal.   Being it a generalization of the problem above, no polynomial time algorithm is known to solve this problem.

 \begin{theorem}
 Let $(\mathcal{P},\Pi)$ be an MC problem with requirements. Let $p$ be a prediction and let $w_i$ be the cost of correcting the $i$th label in $p$.
 For each positive $d$, determining the existence of an admissible prediction $q$ such that $cost(p,q) \le d$ is an NP-complete problem.
  \end{theorem}
This is an easy consequence of Theorem~\ref{th:one}. Luckily, we can again formulate the problem as a PMaxSAT problem in which: 
\begin{enumerate}
    \item each constraint in $\Pi$ corresponds to a clause marked as hard, and
    \item for each $i$ ($1 \le i \le 41$), the prediction in $p$ for the $i$th label corresponds to a unit clause having weight $w_i$. 
\end{enumerate}
Given the above two formulation, we tested two policies: 
\begin{enumerate}
     \item {\sl Average Precision based} (\MSmAP), in which each $w_i$ is equal to the average precision $AP_i$ of the $i$th label, and
    \item {\sl Average Precision and Output based} (\MSPred), in which each $w_i = AP_i \times c_i$, where $c_i$ is equal to (i) the output $o_i$ of the model for the $i$th label if $o_i > \theta$, and (ii) $(1-o_i)$, otherwise.
\end{enumerate}
Differently from the MD policy, in order to take a decision, these policies take into
account the reliability of the output $o_i$ of the model for the $i$th label. 
We again tested the two policies with all the thresholds $\theta$ from 0.1 to 0.9 with step 0.1, 
and the resulting f-mAP@0.5 and f-mAP@0.75 for the best threshold are reported in Table~\ref{tab:map5}, 
columns \MSmAP{}, and \MSPred.
Comparing the predictions of the SOTA models with the CO-\MSmAP{} and the CO-\MSPred{} models, we can see that flipping the variables taking into account the average precision (i) never leads to worse performances than the ones of the  \red{ SOTA models}, and (ii) for IoU = 0.75, \red{correcting the output of RCLSTM with \MSmAP{} and \MSPred{}  gives the best and second best performance in the row.} 
Notice that the differences in the performances between \MSmAP{} and \MSPred{} are always negligible, \MSPred{} being better than \MSmAP{} more often.
\red{The average rankings are in line with the above statements. }

\section{Related Work}
\label{sec:literature}

The approach proposed in this paper generalizes HMC problems, in which requirements are binary and have the form $(A \to B)$, corresponding to our $(\neg A \vee B)$. Many models have been developed for HMC, see e.g., \cite{vens2008,cerri2018,giunchiglia2020}.

Interestingly, when dealing with more complex logical requirements on the output space, researchers have mostly focused on  exploiting the background knowledge that they express to improve performance and/or to deal with data scarcity, curiously neglecting the problem of guaranteeing their satisfaction. Many works go in this direction, such as \cite{hu2016harnessing,hu2016deep}, where an iterative method to embed structured logical information into the neural networks' weights is introduced: at each step, the authors consider a {\sl teacher network}  based on the set of logical rules to train a {\sl student network} to fit both supervisions and logic rules.  
Another neural model is considered in \cite{li2019augmenting}, 
in which some neurons are associated with logical predicates, and their activation is modified on the ground of the activation of the neurons corresponding to predicates that co-occur in the same rules. 

An entire line of research is dedicated to embedding logical constraints into loss functions; see, e.g., \cite{diligenti2017semantic,donadello2017logic,xu2018semantic}. These works consider a fuzzy relaxation of FOL  formulas to get a differentiable loss function that can be 
minimized by gradient descent. However, in all the above methods, there is no guarantee that the constraints will be actually satisfied. 
Luckily recently this problem has gained more relevance, and few works now propose novel ways of addressing the problem. One of such works is \cite{giunchiglia2021},  which presents a novel neural model called {\sl coherent-by-construction network} (CCN). CCN  not only exploits the knowledge expressed by the constraints, but  is also able to guarantee the constraints'  satisfaction. However, that model is able to guarantee the satisfaction of 
 constraints written as 
normal logic rules \red{with at least one positive label}, and thus  is not able to deal with all the ROAD-R's requirements. 

Another work that goes in this direction is  \cite{dragone2021nester}, in which NESTER is proposed. In this case, the constraints are not mapped into the last layer of the  network (like MultiPlexNet or CNN), but they are enforced by passing the outputs of the neural network to a constraint program, which enforces the constraints.
The most recent work is given by \cite{hoernle2022multiplexnet}, where the authors propose  MultiPlexNet. MultiplexNet can impose constraints consisting of any quantifier-free linear arithmetic formula over the rationals (thus, involving ``$+$", ``$\ge$", ``$\neg$", ``$\wedge$", and ``$\vee$"). In order to train the model with such constraints, the formulas are firstly expressed in disjunctive normal form (DNF), and then the output layer of the network is augmented to include a separate transformation for each term in the DNF formula. Thus, the network’s output layer can be viewed as a multiplexor in a logical circuit that permits for a branching of logic. For a general overview of deep learning with logical constraints, see the survey \cite{giunchiglia2022survey}.

In the video understanding field, some recent works have started to argue the importance of being able to extract structured information from videos and to incorporate background knowledge in the models.  For example, \cite{curtis2020} propose a challenge to test the models' ability of extracting knowledge graphs from videos. 
\citeauthor{mahon2020} (\citeyear{mahon2020}) develop a model that is able to exploit the knowledge expressed in logical rules to extract knowledge graphs from videos. However, ROAD-R is the first dataset which proposes the incorporation of logical constraints into deep learning models for videos, and thus represents a truly novel challenge.

\section{Summary and Outlook}
\label{sec:future}

We proposed a new learning framework, called learning with requirements, and a new dataset for this task, called ROAD-R. 
We showed that SOTA models most of the times violate the requirements, and how it is possible to exploit the requirements to create models that are compliant with (i.e., strictly satisfy) the requirements while improving their performance.

We envision that requirement specification will become a standard step in the development of machine learning models, to guarantee their safety, as it is in any software development process. In this sense, ROAD-R may be followed by many other datasets with formally specified requirements.

\section*{Ethical Considerations}

ROAD-R consists of the set of logical constraints in Appendix~\ref{app:req_list}, on top of the existing dataset ROAD, which is publicly available, and which is linked. Thus, ethical issues related to person identification do not apply to ROAD-R. 

\section*{Acknowledgments}
Eleonora Giunchiglia is supported by the EPSRC under the grant EP/N509711/1 and by an Oxford-DeepMind Graduate Scholarship. Mihaela C\u{a}t\u{a}lina Stoian is supported by the EPSRC under the grant EP/T517811/1. Salman Khan is supported by a research agreement with Huawei Technologies Co., Ltd. This work was also supported by the Alan Turing Institute under the EPSRC grant EP/N510129/1, by the AXA Research Fund, by the EPSRC grant EP/R013667/1, and by the European Union’s Horizon 2020 research and innovation programme, under grant agreement No. 964505 (E-pi). We also acknowledge the use of the EPSRC-funded Tier 2 facility JADE (EP/P020275/1) and GPU computing support by Scan Computers International Ltd.

\bibliographystyle{named}
\bibliography{bibliography}

\begin{thebibliography}{}

\bibitem[\protect\citeauthoryear{Bailleux and Marquis}{2006}]{DBLP:BailleuxM06}
Olivier Bailleux and Pierre Marquis.
\newblock Some computational aspects of distance-sat.
\newblock {\em J. Autom. Reason.}, 37(4):231--260, 2006.

\bibitem[\protect\citeauthoryear{Carreira and
  Zisserman}{2017}]{carreira2017quo}
Joao Carreira and Andrew Zisserman.
\newblock Quo vadis, action recognition? {A} new model and the kinetics
  dataset.
\newblock In {\em Proc. of CVPR}, 2017.

\bibitem[\protect\citeauthoryear{Curtis \bgroup \em et al.\egroup
  }{2020}]{curtis2020}
Keith Curtis, George Awad, Shahzad Rajput, and Ian Soboroff.
\newblock {HLVU:} {A} new challenge to test deep understanding of movies the
  way humans do.
\newblock In {\em Proc. of {ICMR}}, 2020.

\bibitem[\protect\citeauthoryear{Demsar}{2006}]{demsar2006}
Janez Demsar.
\newblock Statistical comparisons of classifiers over multiple data sets.
\newblock {\em JMLR}, 7, 2006.

\bibitem[\protect\citeauthoryear{Diligenti \bgroup \em et al.\egroup
  }{2017a}]{diligenti2017semantic}
Michelangelo Diligenti, Marco Gori, and Claudio Sacca.
\newblock Semantic-based regularization for learning and inference.
\newblock {\em Art. Intell.}, 244, 2017.

\bibitem[\protect\citeauthoryear{{Diligenti} \bgroup \em et al.\egroup
  }{2017b}]{diligenti2017icmla}
Michelangelo {Diligenti}, Soumali {Roychowdhury}, and Marco {Gori}.
\newblock Integrating prior knowledge into deep learning.
\newblock In {\em Proc. of ICMLA}, 2017.

\bibitem[\protect\citeauthoryear{Donadello \bgroup \em et al.\egroup
  }{2017}]{donadello2017logic}
Ivan Donadello, Luciano Serafini, and Artur {d'Avila Garcez}.
\newblock Logic tensor networks for semantic image interpretation.
\newblock In {\em Proc. of {IJCAI}}, 2017.

\bibitem[\protect\citeauthoryear{Dragone \bgroup \em et al.\egroup
  }{2021}]{dragone2021nester}
Paolo Dragone, Stefano Teso, and Andrea Passerini.
\newblock Neuro-symbolic constraint programming for structured prediction.
\newblock In {\em {IJCLR-NeSy}}, 2021.

\bibitem[\protect\citeauthoryear{Feichtenhofer \bgroup \em et al.\egroup
  }{2019}]{feichtenhofer2019slowfast}
Christoph Feichtenhofer, Haoqi Fan, Jitendra Malik, and Kaiming He.
\newblock {SlowFast} networks for video recognition.
\newblock In {\em Proc. of ICCV}, 2019.

\bibitem[\protect\citeauthoryear{Giunchigla \bgroup \em et al.\egroup
  }{2022}]{giunchiglia2022survey}
Eleonora Giunchigla, Mihaela~C. Stoian, and Thomas Lukasiewicz.
\newblock Deep learning with logical constraints.
\newblock In {\em Proc. of IJCAI}, 2022.

\bibitem[\protect\citeauthoryear{Giunchiglia and
  Lukasiewicz}{2020}]{giunchiglia2020}
Eleonora Giunchiglia and Thomas Lukasiewicz.
\newblock Coherent hierarchical multi-label classification networks.
\newblock In {\em Proc. of {NeurIPS}}, 2020.

\bibitem[\protect\citeauthoryear{Giunchiglia and
  Lukasiewicz}{2021}]{giunchiglia2021}
Eleonora Giunchiglia and Thomas Lukasiewicz.
\newblock Multi-label classification neural networks with hard logical
  constraints.
\newblock {\em JAIR}, 72, 2021.

\bibitem[\protect\citeauthoryear{He \bgroup \em et al.\egroup
  }{2016}]{he2016deep}
Kaiming He, Xiangyu Zhang, Shaoqing Ren, and Jian Sun.
\newblock Deep residual learning for image recognition.
\newblock In {\em Proc. of CVPR}, 2016.

\bibitem[\protect\citeauthoryear{Hickey and Bacchus}{2019}]{hickey2019}
Randy Hickey and Fahiem Bacchus.
\newblock Speeding up assumption-based {SAT}.
\newblock In {\em Proc. of SAT}, 2019.

\bibitem[\protect\citeauthoryear{Hoernle \bgroup \em et al.\egroup
  }{2022}]{hoernle2022multiplexnet}
Nicholas Hoernle, Rafael{-}Michael Karampatsis, Vaishak Belle, and Kobi Gal.
\newblock {MultiplexNet: T}owards fully satisfied logical constraints in neural
  networks.
\newblock In {\em Proc. of {AAAI}}, 2022.

\bibitem[\protect\citeauthoryear{Hu \bgroup \em et al.\egroup
  }{2016a}]{hu2016harnessing}
Zhiting Hu, Xuezhe Ma, Zhengzhong Liu, Eduard Hovy, and Eric Xing.
\newblock Harnessing deep neural networks with logic rules.
\newblock In {\em Proc. of {ACL}}, 2016.

\bibitem[\protect\citeauthoryear{Hu \bgroup \em et al.\egroup
  }{2016b}]{hu2016deep}
Zhiting Hu, Zichao Yang, Ruslan Salakhutdinov, and Eric Xing.
\newblock Deep neural networks with massive learned knowledge.
\newblock In {\em Proc. of {EMNLP}}, 2016.

\bibitem[\protect\citeauthoryear{Hua \bgroup \em et al.\egroup
  }{2018}]{hua2018traffic}
Yuxiu Hua, Zhifeng Zhao, Zhiming Liu, Xianfu Chen, Rongpeng Li, and Honggang
  Zhang.
\newblock Traffic prediction based on random connectivity in deep learning with
  long short-term memory.
\newblock In {\em Proc. of VTC-Fall}, 2018.

\bibitem[\protect\citeauthoryear{Kalogeiton \bgroup \em et al.\egroup
  }{2017}]{kalogeiton2017a}
Vicky Kalogeiton, Philippe Weinzaepfel, Vittorio Ferrari, and Cordelia Schmid.
\newblock Action tubelet detector for spatio-temporal action localization.
\newblock In {\em Proc. of ICCV}, 2017.

\bibitem[\protect\citeauthoryear{Krizhevsky \bgroup \em et al.\egroup
  }{2012}]{alexnet}
Alex Krizhevsky, Ilya Sutskever, and Geoffrey~E Hinton.
\newblock Imagenet classification with deep convolutional neural networks.
\newblock In {\em Proc. of {NeurIPS}}, 2012.

\bibitem[\protect\citeauthoryear{LeCun \bgroup \em et al.\egroup
  }{2012}]{lecun2012}
Yann LeCun, L{\'{e}}on Bottou, Genevieve Orr, and Klaus{-}Robert M{\"{u}}ller.
\newblock Efficient backprop.
\newblock In {\em Neural Networks: Tricks of the Trade}. 2012.

\bibitem[\protect\citeauthoryear{Li and Many{\`{a}}}{2009}]{DBLP:LiM09}
Chu~Min Li and Felip Many{\`{a}}.
\newblock {MaxSAT}, hard and soft constraints.
\newblock In {\em Handbook of Satisfiability}, volume 185 of {\em Frontiers in
  Artificial Intelligence and Applications}. 2009.

\bibitem[\protect\citeauthoryear{Li and Srikumar}{2019}]{li2019augmenting}
Tao Li and Vivek Srikumar.
\newblock Augmenting neural networks with first-order logic.
\newblock In {\em Proc. of {ACL}}, 2019.

\bibitem[\protect\citeauthoryear{Li \bgroup \em et al.\egroup
  }{2018}]{li2018map}
Dong Li, Zhaofan Qiu, Qi~Dai, Ting Yao, and Tao Mei.
\newblock Recurrent tubelet proposal and recognition networks for action
  detection.
\newblock In {\em Proc. of ECCV}, 2018.

\bibitem[\protect\citeauthoryear{Mahon \bgroup \em et al.\egroup
  }{2020}]{mahon2020}
Louis Mahon, Eleonora Giunchiglia, Bowen Li, and Thomas Lukasiewicz.
\newblock Knowledge graph extraction from videos.
\newblock In {\em Proc. of {ICMLA}}, 2020.

\bibitem[\protect\citeauthoryear{Metcalfe}{2005}]{metcalfe2005}
George Metcalfe.
\newblock Fundamentals of fuzzy logics.
\newblock https://www.logic.at/tbilisi05/Metcalfe-notes.pdf, 2005.

\bibitem[\protect\citeauthoryear{Redmon \bgroup \em et al.\egroup
  }{2016}]{yolo}
Joseph Redmon, Santosh Divvala, Ross Girshick, and Ali Farhadi.
\newblock You only look once: Unified, real-time object detection.
\newblock In {\em Proc. of CVPR}, 2016.

\bibitem[\protect\citeauthoryear{Singh and Cuzzolin}{2019}]{singh2019recurrent}
Gurkirt Singh and Fabio Cuzzolin.
\newblock Recurrent convolutions for causal 3{D CNN}s.
\newblock In {\em Proc. of ICCV Workshops}, 2019.

\bibitem[\protect\citeauthoryear{Singh \bgroup \em et al.\egroup
  }{2021}]{gurkirt2021}
Gurkirt Singh, Stephen Akrigg, Manuele~Di Maio, Valentina Fontana,
  Reza~Javanmard Alitappeh, Suman Saha, Kossar~Jeddi Saravi, Farzad Yousefi,
  Jacob Culley, Tom Nicholson, Jordan Omokeowa, Salman Khan, Stanislao
  Grazioso, Andrew Bradley, Giuseppe~Di Gironimo, and Fabio Cuzzolin.
\newblock {ROAD: T}he road event awareness dataset for autonomous driving.
\newblock {\em IEEE TPAMI}, 2021.

\bibitem[\protect\citeauthoryear{Sommerville}{2011}]{sommerville}
Ian Sommerville.
\newblock {\em {Software Engineering}}.
\newblock 2011.

\bibitem[\protect\citeauthoryear{Vens \bgroup \em et al.\egroup
  }{2008}]{vens2008}
Celine Vens, Jan Struyf, Leander Schietgat, Saso Dzeroski, and Hendrik
  Blockeel.
\newblock Decision trees for hierarchical multi-label classification.
\newblock {\em Mach. Learn.}, 73, 2008.

\bibitem[\protect\citeauthoryear{Wang \bgroup \em et al.\egroup
  }{2018}]{wang2018non}
Xiaolong Wang, Ross Girshick, Abhinav Gupta, and Kaiming He.
\newblock Non-local neural networks.
\newblock In {\em Proc. of CVPR}, 2018.

\bibitem[\protect\citeauthoryear{Wehrmann \bgroup \em et al.\egroup
  }{2018}]{cerri2018}
Jonatas Wehrmann, Ricardo Cerri, and Rodrigo~C. Barros.
\newblock Hierarchical multi-label classification networks.
\newblock In {\em Proc. of {ICML}}, 2018.

\bibitem[\protect\citeauthoryear{Xu \bgroup \em et al.\egroup
  }{2018}]{xu2018semantic}
Jingyi Xu, Zilu Zhang, Tal Friedman, Yitao Liang, and Guy Van~den Broeck.
\newblock A semantic loss function for deep learning with symbolic knowledge.
\newblock In {\em Proc. of {ICML}}, 2018.

\end{thebibliography}

\clearpage 

\appendix

\section{ROAD Labels}
\label{app:lab_description}

Here, we provide a detailed description of the meaning of each of the 41 labels. In particular, we describe
\begin{enumerate}
    \item the labels associated with agents in Table~\ref{tab:agent}, 
    \item  the labels associated with actions in Table~\ref{tab:action},  and 
    \item the labels associated with locations  in Table~\ref{tab:location}.
\end{enumerate}
In the last column of each table, we also report the abbreviations used in our implementation and associated with each label. Such abbreviations were taken from \url{https://github.com/gurkirt/3D-RetinaNet}, and allowed a seamless integration of our code with the SOTA models' code. 
The abbreviations are used in Tables \ref{tab:req_list1}, \ref{tab:req_list2}, and \ref{tab:req_list3} to write the requirements using the set based notation.

\section{Requirements}\label{app:req_list}

In Tables~\ref{tab:req_list1}, \ref{tab:req_list2}, and \ref{tab:req_list3}, we report the list of all the 243 requirements together with their natural language explanations.

\section{Visual Examples of Violations}\label{app:imgs}

\begin{figure*}

\begin{subfigure}[b]{0.33\textwidth}
         \centering
         \includegraphics[width=\textwidth]{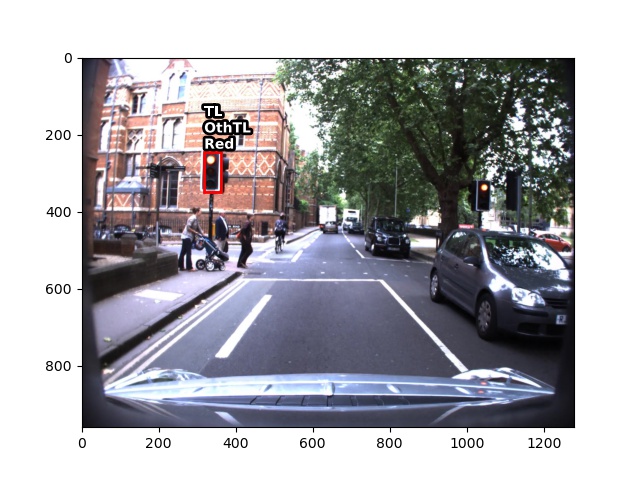}
         \caption{I3D}
         \label{I3D2}
\end{subfigure}
\begin{subfigure}[b]{0.33\textwidth}
         \centering
         \includegraphics[width=\textwidth]{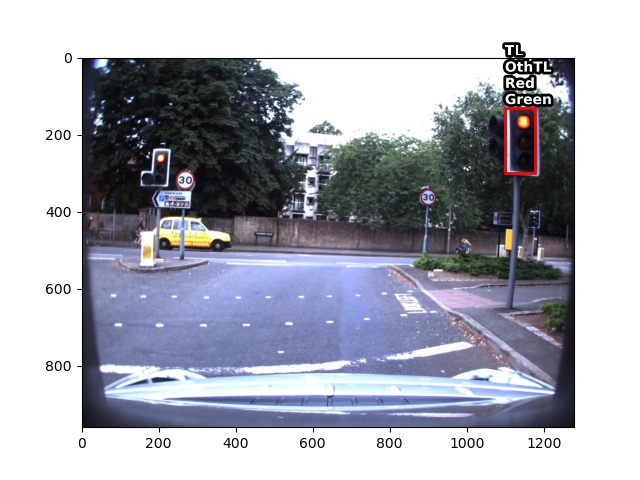}
         \caption{C2D}
         \label{C2D2}
\end{subfigure}
\begin{subfigure}[b]{0.33\textwidth}
         \centering
         \includegraphics[width=\textwidth]{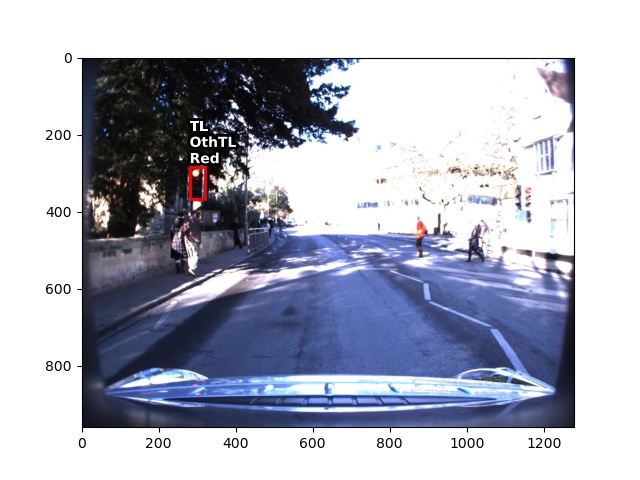}
         \caption{RCGRU}
         \label{RCGRU2}
\end{subfigure}

\begin{subfigure}[b]{0.33\textwidth}
         \centering
         \includegraphics[width=\textwidth]{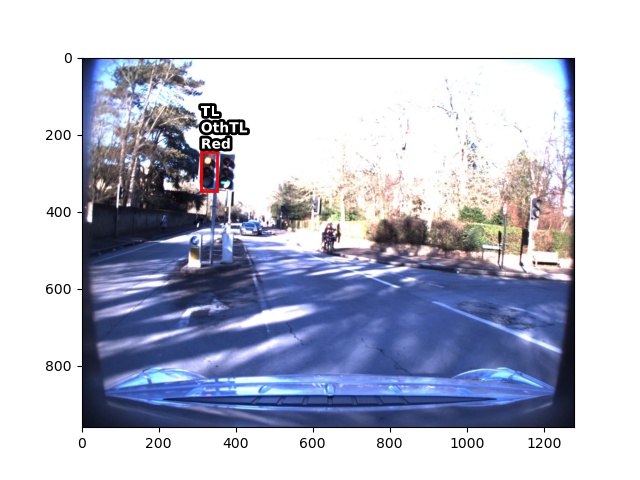}
         \caption{RCLSTM}
         \label{RCLSTM2}
\end{subfigure}
\begin{subfigure}[b]{0.33\textwidth}
         \centering
         \includegraphics[width=\textwidth]{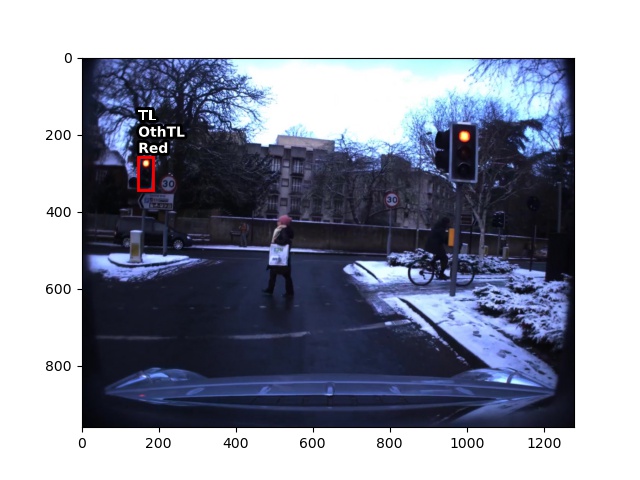}
         \caption{RCN}
         \label{RCN2}
\end{subfigure}
\begin{subfigure}[b]{0.33\textwidth}
         \centering
         \includegraphics[width=\textwidth]{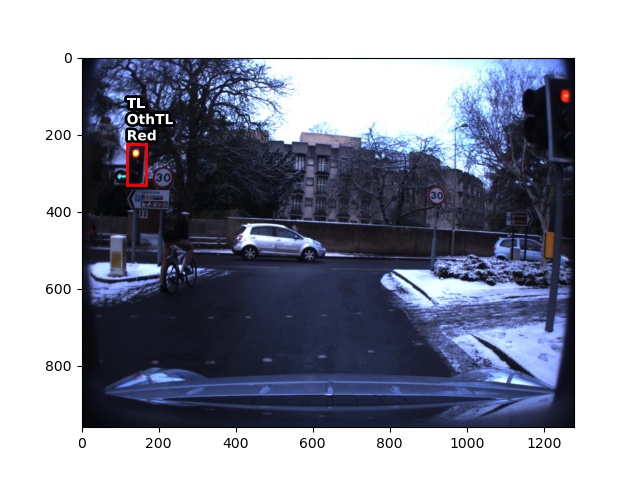}
         \caption{SlowFast}
         \label{SLOWFAST2}
\end{subfigure}
\caption{Examples of violations of $\{\neg{\text{TL}},\neg{\text{OthTL}}\}$.}
\label{fig:violOthTL}
\end{figure*}

\begin{figure*}

\begin{subfigure}[b]{0.33\textwidth}
         \centering
         \includegraphics[width=\textwidth]{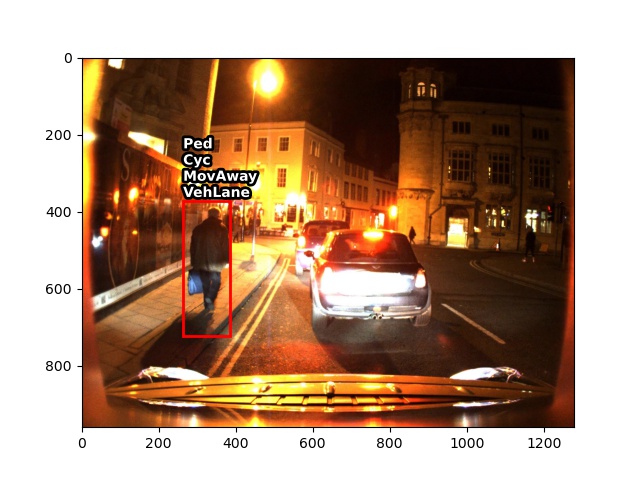}
         \caption{I3D}
         \label{I3D3}
\end{subfigure}
\begin{subfigure}[b]{0.33\textwidth}
         \centering
         \includegraphics[width=\textwidth]{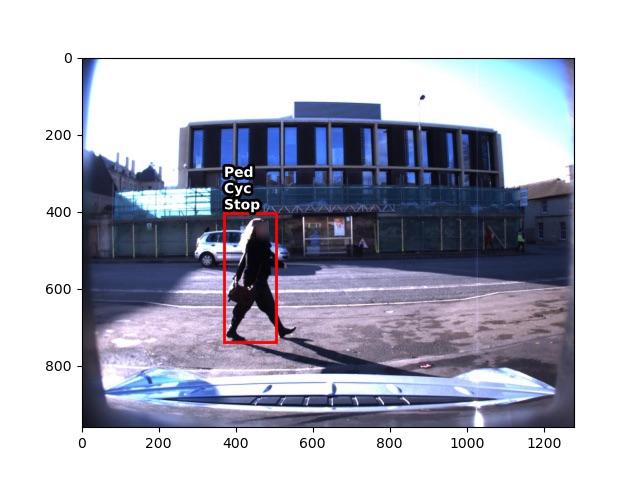}
         \caption{C2D}
         \label{C2D3}
\end{subfigure}
\begin{subfigure}[b]{0.33\textwidth}
         \centering
         \includegraphics[width=\textwidth]{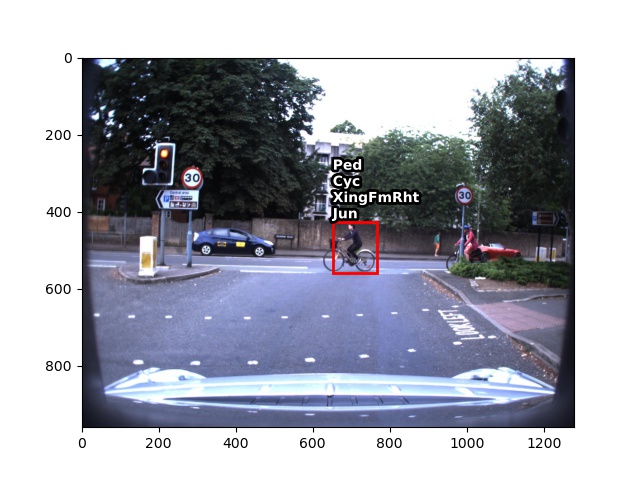}
         \caption{RCGRU}
         \label{RCGRU3}
\end{subfigure}

\begin{subfigure}[b]{0.33\textwidth}
         \centering
         \includegraphics[width=\textwidth]{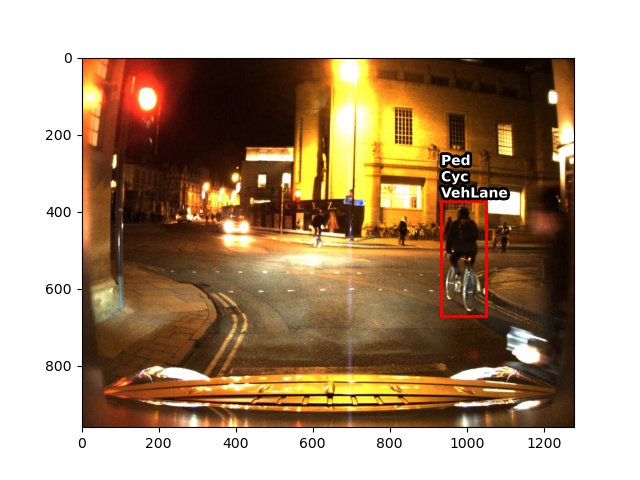}
         \caption{RCLSTM}
         \label{RCLSTM3}
\end{subfigure}
\begin{subfigure}[b]{0.33\textwidth}
         \centering
         \includegraphics[width=\textwidth]{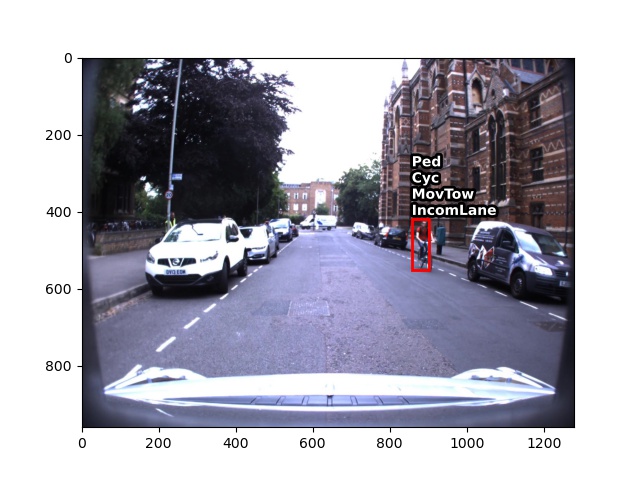}
         \caption{RCN}
         \label{RCN3}
\end{subfigure}
\begin{subfigure}[b]{0.33\textwidth}
         \centering
         \includegraphics[width=\textwidth]{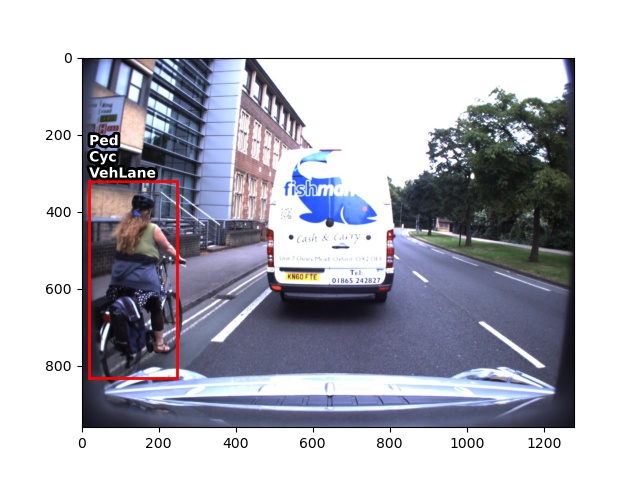}
         \caption{SlowFast}
\label{SLOWFAST3}
\end{subfigure}

\caption{Examples of violations of $\{\neg{\text{Ped}},\neg{\text{Cyc}}\}$.}
\label{fig:violPed}
\end{figure*}

\begin{figure*}

\begin{subfigure}[b]{0.33\textwidth}
         \centering
         \includegraphics[width=\textwidth]{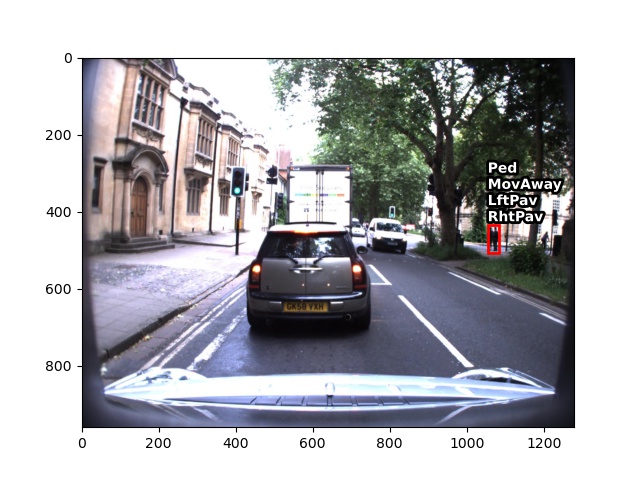}
         \caption{I3D}
         \label{I3D4}
\end{subfigure}
\begin{subfigure}[b]{0.33\textwidth}
         \centering
         \includegraphics[width=\textwidth]{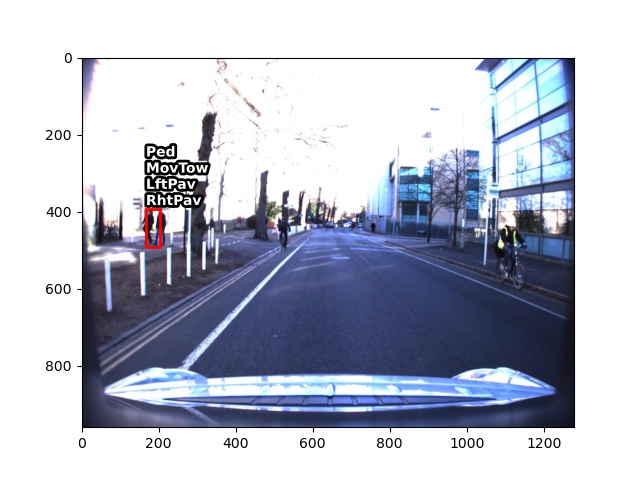}
         \caption{C2D}
         \label{C2D4}
\end{subfigure}
\begin{subfigure}[b]{0.33\textwidth}
         \centering
         \includegraphics[width=\textwidth]{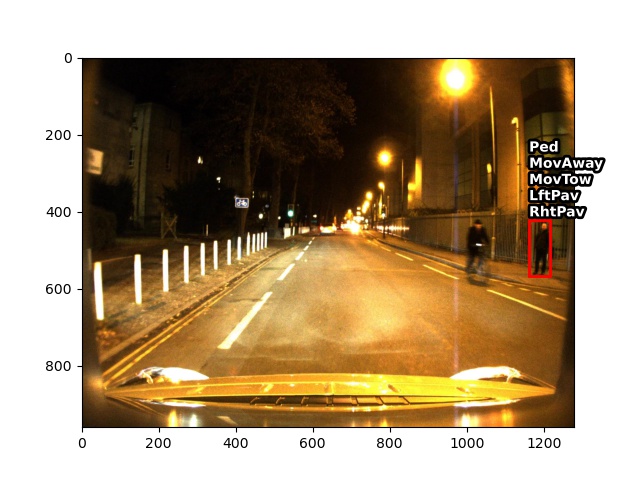}
         \caption{RCGRU}
         \label{RCGRU4}
\end{subfigure}

\begin{subfigure}[b]{0.33\textwidth}
         \centering
         \includegraphics[width=\textwidth]{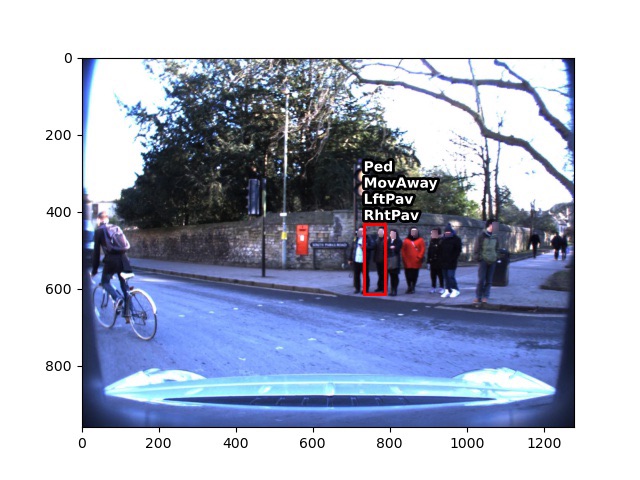}
         \caption{RCLSTM}
         \label{RCLSTM4}
\end{subfigure}
\begin{subfigure}[b]{0.33\textwidth}
         \centering
         \includegraphics[width=\textwidth]{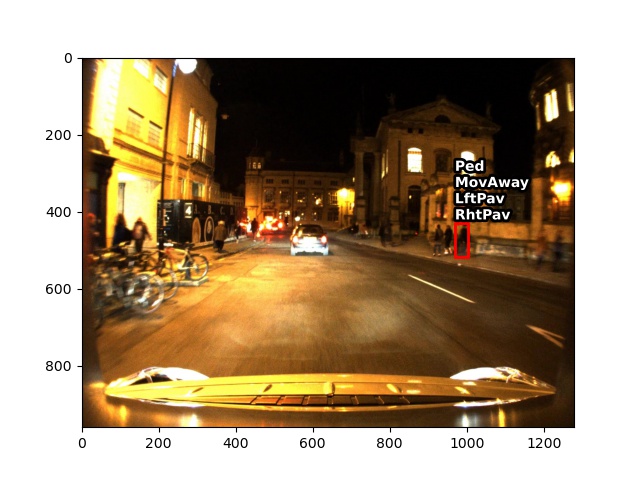}
         \caption{RCN}
         \label{RCN4}
\end{subfigure}
\begin{subfigure}[b]{0.33\textwidth}
         \centering
         \includegraphics[width=\textwidth]{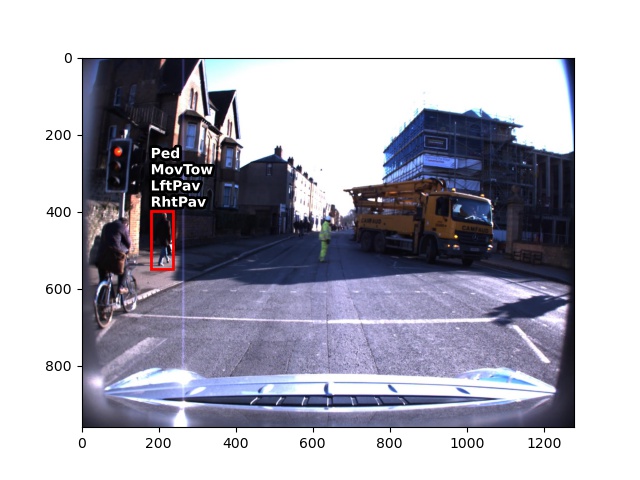}
         \caption{SlowFast}
         \label{SLOWFAST4}
\end{subfigure}
\caption{Examples of violations of $\{\neg{\text{LeftPav}},\neg{\text{RightPav}}\}$.}
\label{fig:violPav}
\end{figure*}

\begin{figure*}

\begin{subfigure}[b]{0.33\textwidth}
         \centering
         \includegraphics[width=\textwidth]{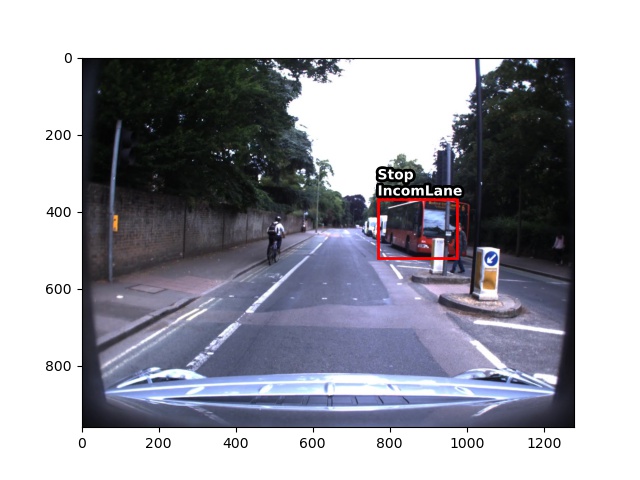}
         \caption{I3D}
         \label{I3D5}
\end{subfigure}
\begin{subfigure}[b]{0.33\textwidth}
         \centering
         \includegraphics[width=\textwidth]{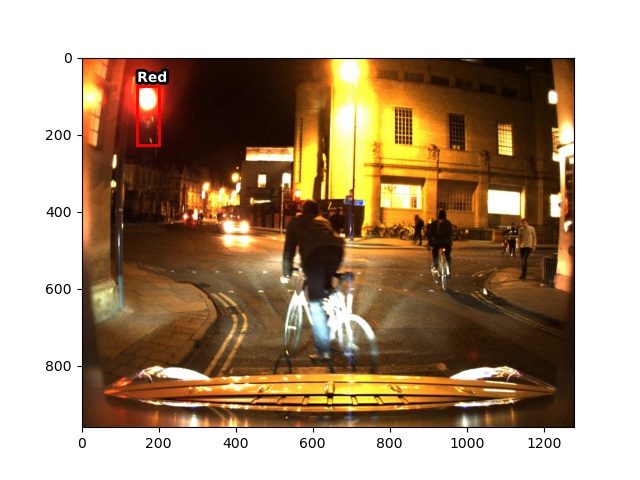}
         \caption{C2D}
         \label{C2D5}
\end{subfigure}
\begin{subfigure}[b]{0.33\textwidth}
         \centering
         \includegraphics[width=\textwidth]{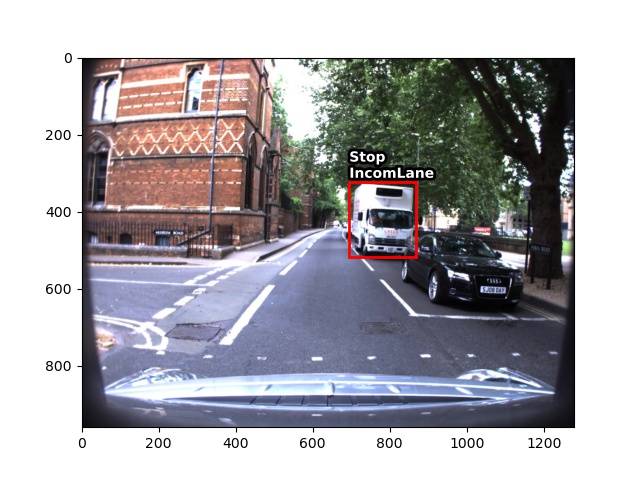}
         \caption{RCGRU}
         \label{RCGRU5}
\end{subfigure}

\begin{subfigure}[b]{0.33\textwidth}
         \centering
         \includegraphics[width=\textwidth]{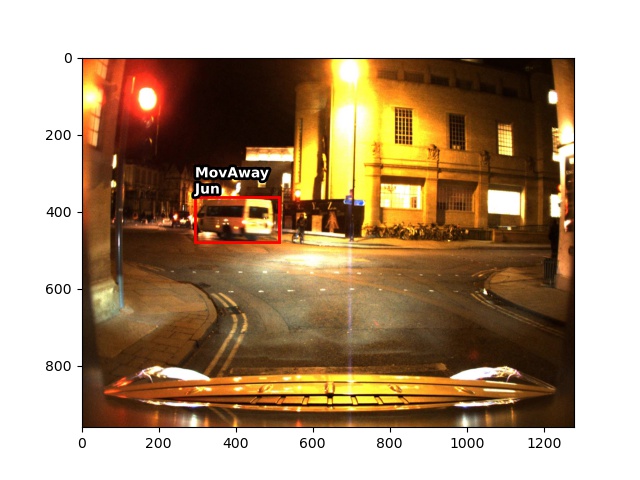}
         \caption{RCLSTM}
         \label{RCLSTM5}
\end{subfigure}
\begin{subfigure}[b]{0.33\textwidth}
         \centering
         \includegraphics[width=\textwidth]{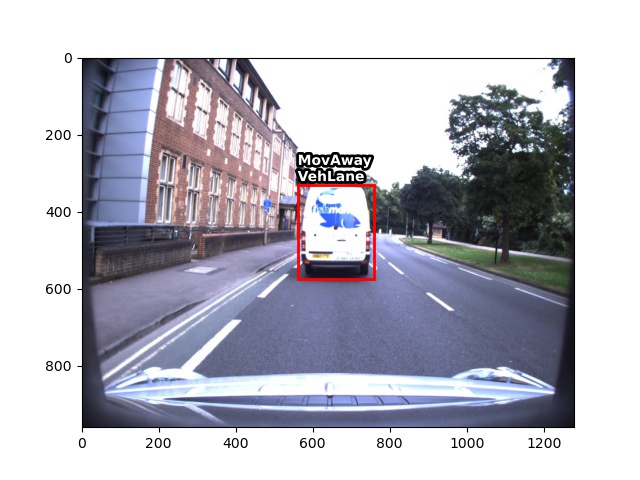}
         \caption{RCN}
         \label{RCN5}
\end{subfigure}
\begin{subfigure}[b]{0.33\textwidth}
         \centering
         \includegraphics[width=\textwidth]{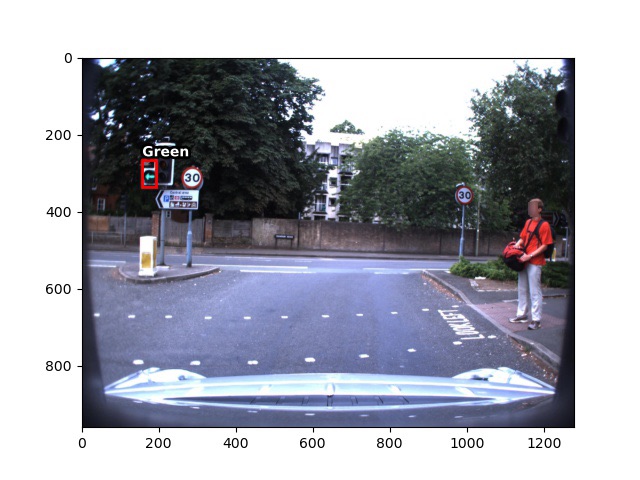}
         \caption{RCN}
         \label{SLOWFAST5}
\end{subfigure}
\caption{Examples of violations of \{Ped, Car, Cyc, Mobike, MedVeh, LarVeh, Bus, EmVeh, TL, OthTL\}.}
\label{fig:violAtLeastOne}
\end{figure*}

\begin{figure*}

\begin{subfigure}[b]{0.33\textwidth}
         \centering
         \includegraphics[width=\textwidth]{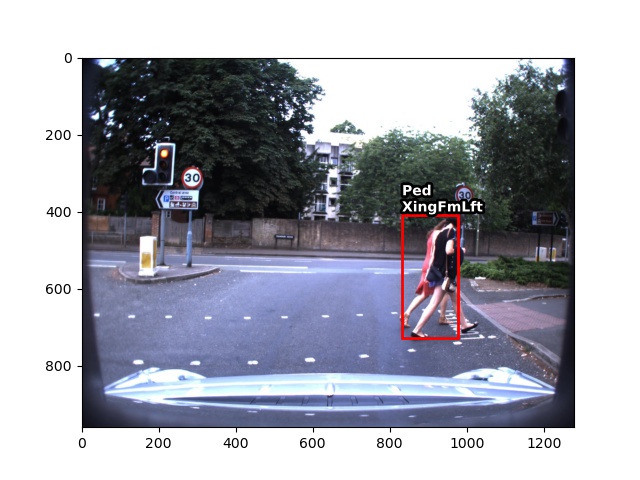}
         \caption{I3D}
         \label{I3D6}
\end{subfigure}
\begin{subfigure}[b]{0.33\textwidth}
         \centering
         \includegraphics[width=\textwidth]{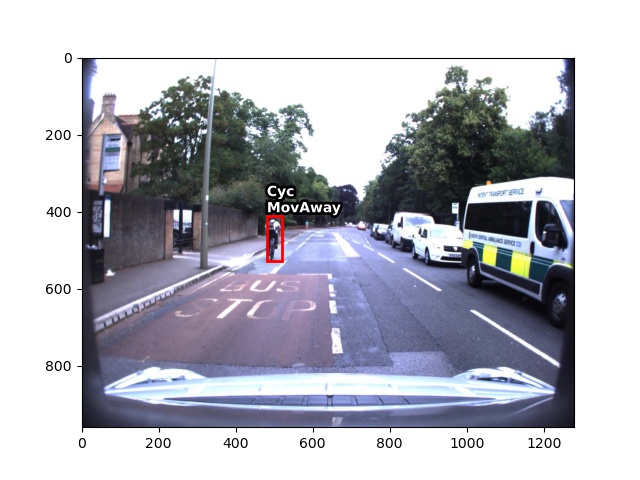}
         \caption{C2D}
         \label{C2D6}
\end{subfigure}
\begin{subfigure}[b]{0.33\textwidth}
         \centering
         \includegraphics[width=\textwidth]{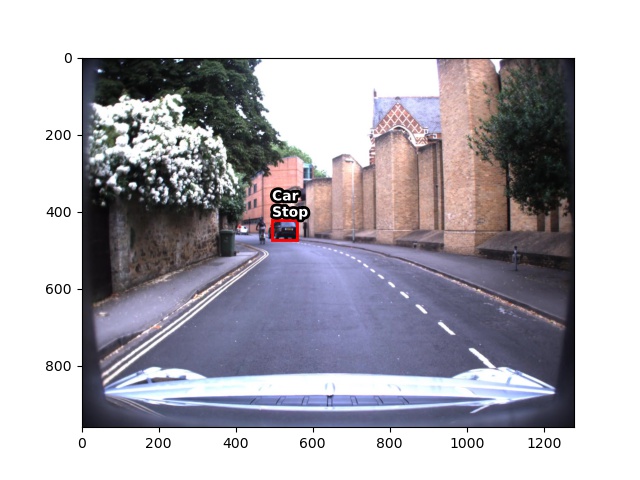}
         \caption{RCGRU}
         \label{RCGRU6}
\end{subfigure}

\hfill
\begin{subfigure}[b]{0.33\textwidth}
         \centering
         \includegraphics[width=\textwidth]{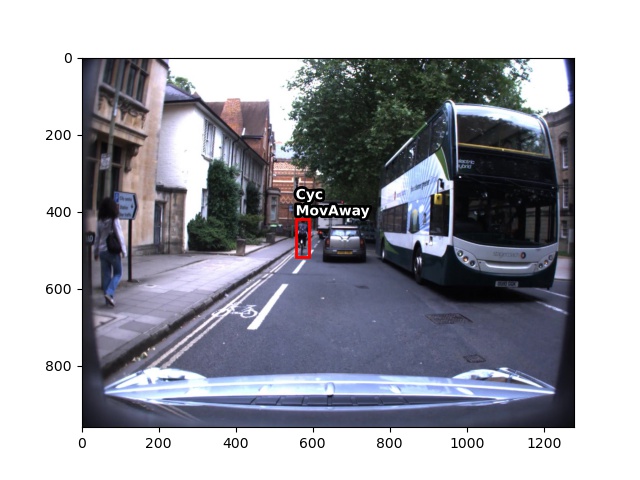}
         \caption{RCLSTM}
         \label{RCLSTM6}
\end{subfigure}
\begin{subfigure}[b]{0.33\textwidth}
         \centering
         \includegraphics[width=\textwidth]{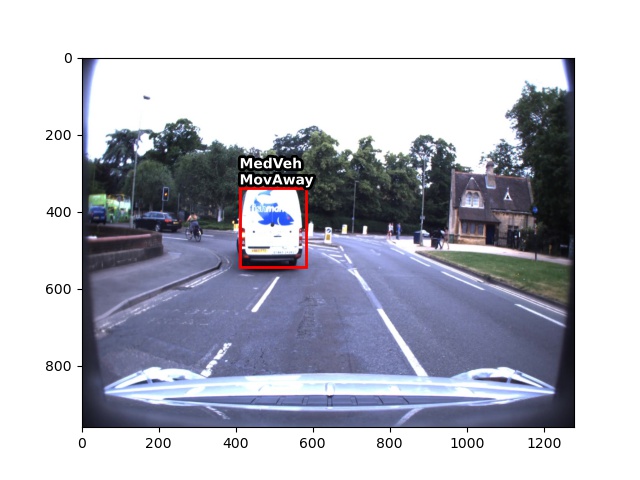}
         \caption{RCN}
         \label{RCN6}
\end{subfigure}
\begin{subfigure}[b]{0.33\textwidth}
         \centering
         \includegraphics[width=\textwidth]{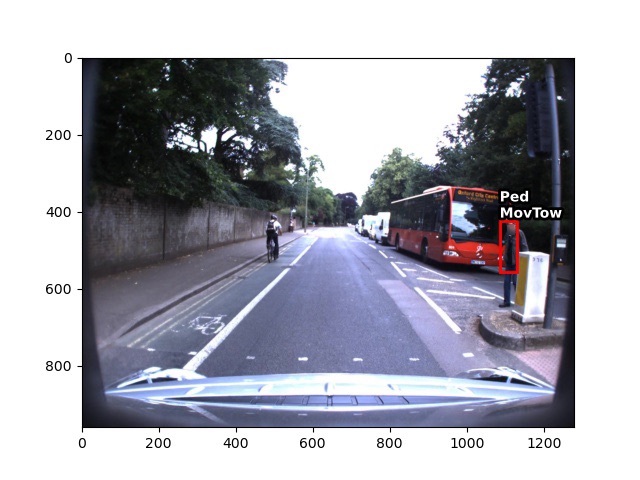}
         \caption{SlowFast}
         \label{SLOWFAST6}
\end{subfigure}
\caption{Examples of violations of
     \{TL, OthTL, VehLane, OutgoLane, OutgoCycLane, Jun,  IncomLane, IncomCycLane, Pav, LftPav, RhtPav, XingLoc, BusStop, Parking\}.}
\label{fig:viol_atleastoneloc}
\end{figure*}

\begin{figure*}

\begin{subfigure}[b]{0.33\textwidth}
         \centering
         \includegraphics[width=\textwidth]{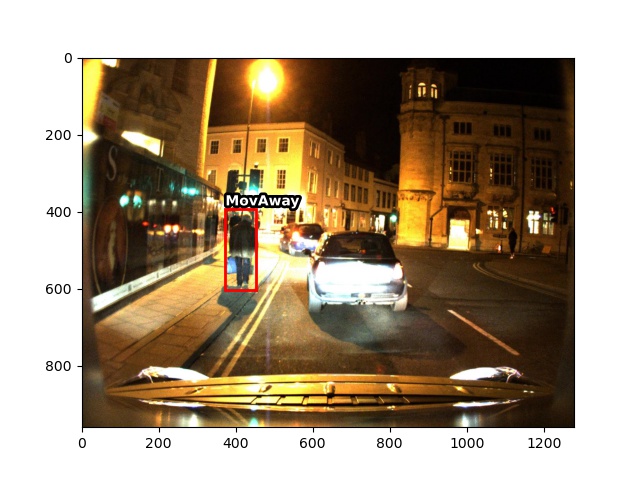}
         \caption{I3D}
         \label{I3D7}
\end{subfigure}
\begin{subfigure}[b]{0.33\textwidth}
         \centering
         \includegraphics[width=\textwidth]{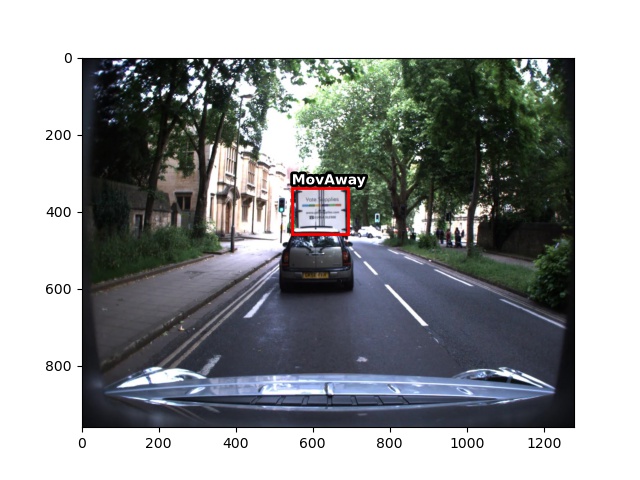}
         \caption{C2D}
         \label{C2D7}
\end{subfigure}
\begin{subfigure}[b]{0.33\textwidth}
         \centering
         \includegraphics[width=\textwidth]{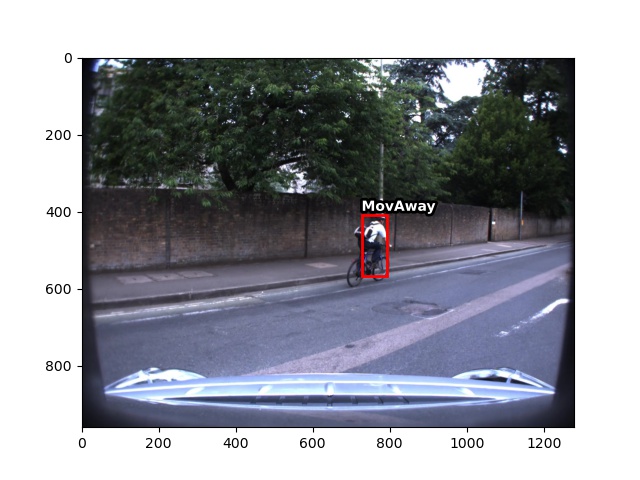}
         \caption{RCGRU}
         \label{RCGRU7}
\end{subfigure}

\begin{subfigure}[b]{0.33\textwidth}
         \centering
         \includegraphics[width=\textwidth]{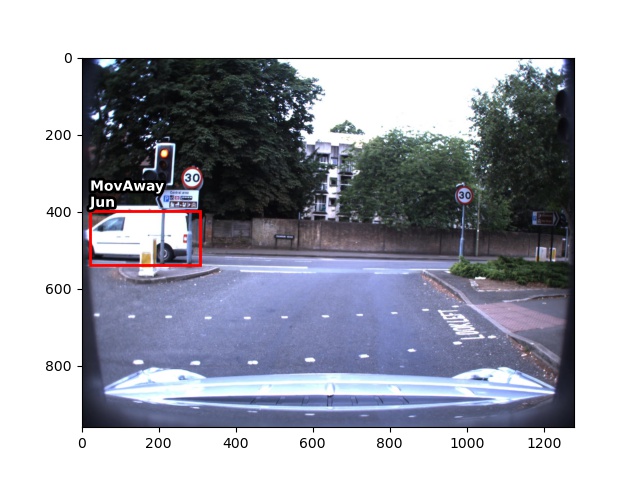}
         \caption{RCLSTM}
         \label{RCLSTM7}
\end{subfigure}
\begin{subfigure}[b]{0.33\textwidth}
         \centering
         \includegraphics[width=\textwidth]{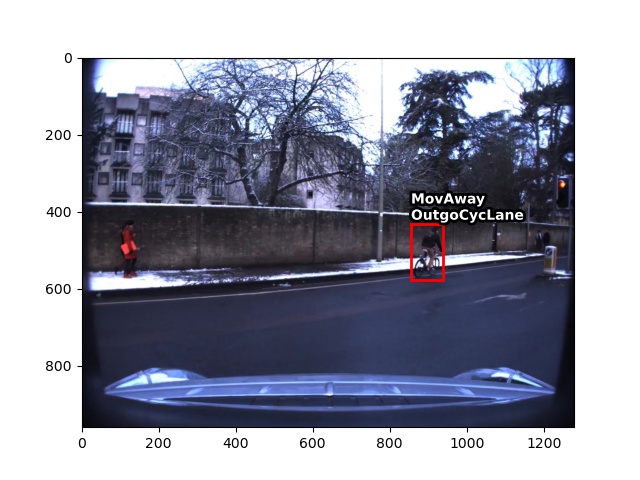}
         \caption{RCN}
         \label{RCN7}
\end{subfigure}
\begin{subfigure}[b]{0.33\textwidth}
         \centering
         \includegraphics[width=\textwidth]{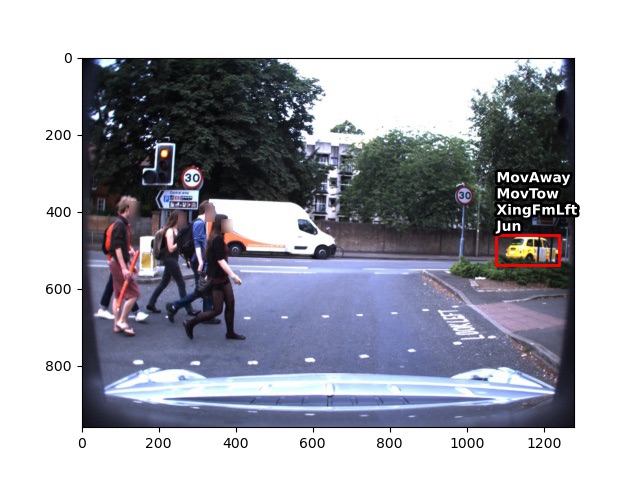}
         \caption{SlowFast}
         \label{SLOWFAST7}
\end{subfigure}
\hfill
\hfill
\caption{Examples of violations of
     \{Ped, Car, Cyc, Mobike, MedVeh, LarVeh, Bus, EmVeh, $\neg{\text{MovAway}}$\}.}
\label{fig:viol_movaway}
\end{figure*}

In this section, we give some visual examples of violations. 
In order to show different situations in which a non-admissible prediction occurs, we consider various requirements, and (for each  of them) we display a non-admissible prediction made by each SOTA model. For all examples, we pick the threshold $\theta = 0.5$, as it is the most intuitive and used threshold in multi-label classification problems. 
In particular, we show the violations for: 
\begin{enumerate}
    \item $\{\neg {\text{TL}},\neg {\text{OthTL}}\}$ in Figure~\ref{fig:violOthTL} ,
    \item $\{\neg{\text{Ped}},\neg{\text{Cyc}}\}$ in Figure~\ref{fig:violPed},
    \item $\{\neg{\text{LeftPav}},\neg{\text{RightPav}}\}$ in Figure~\ref{fig:violPav}, 
    \item \{Ped, Car, Cyc, Mobike, MedVeh, LarVeh, Bus, EmVeh, TL, OthTL\} in Figure~\ref{fig:violAtLeastOne},
    \item \{TL, OthTL, VehLane, OutgoLane,OutgoCycLane, Jun  IncomLane, IncomCycLane, Pav, LftPav, RhtPav, XingLoc, BusStop, Parking\} in Figure~\ref{fig:viol_atleastoneloc}, and 
    \item \{Ped, Car, Cyc, Mobike, MedVeh, LarVeh, Bus, EmVeh, $\neg{\text{MovAway}}$\} in Figure~\ref{fig:viol_movaway}.
\end{enumerate}
In the above list, there is at least one requirement with (i) all negative labels, (ii) all positive labels, and (iii) at least one positive and one negative label. Furthermore, at least one agent label, one action label, and one location label appear in at least one of the above requirements.

\begin{table*}[t]
\footnotesize
\centering
\begin{tabular}{l l l}
\toprule 
\textbf{Label name} & \textbf{Description} & \textbf{Abbrv.}
\\
\midrule 
Pedestrian & A person including children & Ped
\\
Car & A car up to the size of a multi-purpose vehicle & Car
\\
Cyclist & A person is riding a push/electric bicycle & Cyc
\\
Motorbike & Motorbike, dirt bike, scooter with 2/3 wheels & Mobike
\\
Medium vehicle & Vehicle larger than a car, such as van & MedVeh
\\
Large vehicle & Vehicle larger than a van, such as a lorry & LarVeh 
\\
Bus & A single or double-decker bus or coach & Bus
\\
Emergency vehicle & Ambulance, police car, fire engine, etc. & EmVeh
\\
AV traffic light & Traffic light related to the AV lane & TL
\\
Other traffic light & Traffic light not related to the AV lane & OthTL
\\
\bottomrule
\end{tabular}
\caption{Agent labels with descriptions and abbreviations \protect \cite{gurkirt2021}. }
\label{tab:agent}
\end{table*}

\begin{table*}[t!]
\footnotesize
  \centering
  \begin{tabular}{l l l}
  \toprule
  \textbf{Label name} & \textbf{Description} & \textbf{Abbrv.}
  \\
  \midrule
  Move away & Agent moving in a direction that increases the distance between Agent and AV & MovAway 
  \\
  Move towards & Agent moving in a direction that decreases the distance between Agent and AV & MovTow
  \\
  Move & Agent moving perpendicular to the traffic flow or vehicle lane & Mov
  \\
   Brake & Agent is slowing down, vehicle braking lights are lit & Brake
  \\
    Stop & Agent stationary but in ready position to move & Stop 
  \\
    Indicating left & Agent indicating left by flashing left indicator light, or using a hand signal & IncatLeft
  \\
  Indicating right & Agent indicating right by flashing right indicator light, or using a hand signal & IncatRht
  \\
    Hazard lights on & Hazards lights are flashing on a vehicle & HazLit
  \\
    Turn left & Agent is turning in left direction & TurLft
  \\
  Turn right & Agent is turning in right direction & TurRht
  \\
    Overtake & Agent is moving around a slow-moving user, often switching lanes to overtake & Ovtak
  \\
    Wait to cross & Agent on a pavement, stationary, facing in the direction of the road & Wait2X
  \\
    Cross road from left & Agent crossing road, starting from the left and moving towards the right of AV & XingFmLft
  \\
  Cross road from right & Agent crossing road, starting from the right and moving towards the left of AV & XingFmRht
  \\
  Crossing & Agent crossing road & Xing
  \\
    Push object & Agent pushing object, such as trolley or pushchair, wheelchair or bicycle & PushObj
    \\
  Red traffic light & Traffic light with red light lit & Red
  \\
  Amber traffic light & Traffic light with amber light lit & Amber
  \\
  Green traffic light & Traffic light with green light lit & Green 
  \\
  \bottomrule
  \end{tabular}
    \caption{Action labels with descriptions and abbreviations \protect \cite{gurkirt2021}. }
    \label{tab:action}
  \end{table*}
  \begin{table*}[t]
  \footnotesize
    \centering
    \begin{tabular}{l l l}
    \toprule
    \textbf{Label name} & \textbf{Description} & \textbf{Abbrv.} \\
    \midrule
    AV lane & Agent in same road lane as AV & VehLane\\
    Outgoing lane & Agent in road lane that should be flowing in the same direction as vehicle lane & OutgoLane \\
    Outgoing cycle lane & Agent in the cycle lane that should be flowing in the same direction as AV & OutgoCycLane \\
    Incoming lane & Agent in road lane that should be flowing in the opposite direction as vehicle lane & IncomLane \\
    Incoming cycle lane & Agent in the cycle lane that should be flowing in the opposite direction as AV & IncomCycLane \\
    Pavement & A pavement that is perpendicular to the movement of the AV & Pav\\ 
    Left pavement & Pavement to the left side of AV & LftPav \\
    Right pavement & Pavement to the right side of AV & RhtPav \\
    Junction & Road linked & Jun \\
    Crossing location & A marked section of road for cross, such as zebra or pelican crossing & XingLoc\\
    Bus stop & A marked bus stop area on road, or a section of pavement next to a bus stop sign & BusStop \\
    Parking & A marked parking area on the side of the road & Parking\\
    \bottomrule
    \end{tabular}
    \caption{Location labels with descriptions and abbreviations \protect \cite{gurkirt2021}.}
    \label{tab:location}
    \end{table*}  

\begin{landscape}
\begin{table}[t]
    \scriptsize
    \centering
    \begin{tabular}{l l }
        \toprule
        \textbf{Requirements} & \textbf{Natural Language Explanations}\\
        \midrule
         \{Ped, not PushObj\} & If an agent pushes an object then it is a pedestrian \\
         \{PushObj, not Ped, MovAway, MovTow, Mov, Stop, TurLft, TurRht, Wait2X, XingFmLft, XingFmRht, Xing\} & A pedestrian can only push objects, move away, etc. \\                                      
\{Ped, not XingFmLft, Car, Cyc, Mobike, MedVeh, LarVeh, Bus, EmVeh\}                                                              &  Only pedestrains, cars, cyclists, etc. can cross from left \\
\{Ped, not Wait2X, Cyc\}                                                                                                          &  Only pedestrians and cyclists can wait to cross \\
\{Ped, not Stop, Car, Cyc, Mobike, MedVeh, LarVeh, Bus, EmVeh\}                                                                   & Only pedestrians, cars, cyclists, etc can stop \\
\{Ped, not Mov, Car, Cyc, Mobike, MedVeh, LarVeh, Bus, EmVeh\}                                                                    & Only pedestrians, cars, cyclists, etc can move \\
\{Ped, not MovTow, Car, Cyc, Mobike, MedVeh, LarVeh, Bus, EmVeh\}                                                                 &  Only pedestrians, cars, cyclists, etc can move towards \\
\{Ped, not MovAway, Car, Cyc, Mobike, MedVeh, LarVeh, Bus, EmVeh\}                                                                & Only pedestrians, cars, cyclists, etc can move away  \\
\{Ovtak, not EmVeh, MovAway, MovTow, Mov, Brake, Stop, IncatLeft, IncatRht, HazLit, TurLft, TurRht, XingFmRht, XingFmLft, Xing\}  &  An emergency vehicle can only overtake, move away  etc. \\
\{EmVeh, not HazLit, Car, MedVeh, LarVeh, Bus, Mobike\}                                                                           & Only emergency vehicles, cars etc. can have hazards lights on \\
\{Ovtak, not Bus, MovAway, MovTow, Mov, Brake, Stop, IncatLeft, IncatRht, HazLit, TurLft, TurRht, XingFmRht, XingFmLft, Xing\}    & A bus can only overtake, move away move towards etc. \\
\{Ovtak, not MedVeh, MovAway, MovTow, Mov, Brake, Stop, IncatLeft, IncatRht, HazLit, TurLft, TurRht, XingFmRht, XingFmLft, Xing\} & A medium vehicle can only overtake, move away, move towards etc. \\
\{Ovtak, not LarVeh, MovAway, MovTow, Mov, Brake, Stop, IncatLeft, IncatRht, HazLit, TurLft, TurRht, XingFmRht, XingFmLft, Xing\} & A large vehicle can only overtake, move away, move towards etc. \\
\{OthTL, not Green, TL\}                                                                                                          &  Only traffic lights and other traffic lights can give a green signal\\
\{OthTL, not Amber, TL\}                                                                                                          &  Only traffic lights and other traffic lights can give an amber signal \\
\{OthTL, not Red, TL\}                                                                                                            & Only traffic lights and other traffic lights can give a red signal \\
\{Ovtak, not Mobike, MovAway, MovTow, Mov, Brake, Stop, IncatLeft, IncatRht, HazLit, TurLft, TurRht, XingFmRht, XingFmLft, Xing\} &  A motorbike can only overtake, move away, move towards etc.\\
\{Xing, not Cyc, MovAway, MovTow, Mov, Brake, Stop, IncatLeft, IncatRht, TurLft, TurRht, Ovtak, Wait2X, XingFmLft, XingFmRht\}    & A cyclist can only cross, move away, move towards etc.  \\
\{Cyc, not Ovtak, MedVeh, LarVeh, Bus, Mobike, EmVeh, Car\}                                                                       &  Only cyclists, medium vehicles, large vehicles etc. can overtake\\
\{Cyc, not IncatRht, MedVeh, LarVeh, Bus, Mobike, EmVeh, Car\}                                                                    & Only cyclists, medium vehicles, large vehicles etc. can indicate right \\
\{Cyc, not IncatLeft, MedVeh, LarVeh, Bus, Mobike, EmVeh, Car\}                                                                   & Only cyclists, medium vehicles, large vehicles etc. can indicate left \\
\{Cyc, not Brake, MedVeh, LarVeh, Bus, Mobike, EmVeh, Car\}                                                                       & Only cyclists, medium vehicles, large vehicles etc. can brake \\
\{Ovtak, not Car, MovAway, MovTow, Mov, Brake, Stop, IncatLeft, IncatRht, HazLit, TurLft, TurRht, XingFmRht, XingFmLft, Xing\}    & A car can only overtake, move away, move towards etc. \\
\{Car, not TurRht, Cyc, Mobike, MedVeh, LarVeh, Bus, EmVeh\}                                                                      & Only cyclists, medium vehicles, large vehicles etc. can turn right \\
\{Car, not TurLft, Cyc, Mobike, MedVeh, LarVeh, Bus, EmVeh\}                                                                      &  Only cyclists, medium vehicles, large vehicles etc. can turn left \\
\{VehLane, OutgoLane, OutgoCycLane, IncomLane, IncomCycLane, Pav, LftPav, RhtPav, Jun, XingLoc, BusStop, Parking, TL, OthTL\} & Every agent but traffic lights must have a position \\
\{Ped, Car, Cyc, Mobike, MedVeh, LarVeh, Bus, EmVeh, TL, OthTL\} & There must be at least an agent \\
\{not Car, not Cyc\}                                                                                                              & A car cannot be a cyclist \\
\{not Car, not Mobike\}                                                                                                           & A car cannot be a motorbike \\
\{not Car, not MedVeh\}                                                                                                           & A car cannot be a medium vehicle \\
\{not Car, not LarVeh\}                                                                                                           & A car cannot be a large vehicle \\
\{not Car, not Bus\}                                                                                                              & A car cannot be a bus \\
\{not Cyc, not Mobike\}                                                                                                           &  A cyclist cannot be a motorbike\\
\{not Car, not EmVeh\}                                                                                                            &  A car cannot be a emergency vehicle \\
\{not Car, not TL\}                                                                                                               & A car cannot be a traffic light\\
\{not Cyc, not MedVeh\}                                                                                                           & A cyclist cannot be a  medium vehicle\\
\{not Car, not OthTL\}                                                                                                            & A car cannot be a other traffic light\\
\{not Car, not Red\}                                                                                                              & A car cannot signal red\\
\{not Cyc, not LarVeh\}                                                                                                           & A cyclist cannot be a large vehicle\\
\{not Car, not Amber\}                                                                                                            & A car cannot signal amber \\
\{not Car, not Green\}                                                                                                            &  A car cannot signal green \\
\{not Cyc, not Bus\}                                                                                                              & A cyclist cannot be a bus \\
\{not Mobike, not MedVeh \} & A motorbike cannot be a medium vehicle \\  %
\{not Cyc, not EmVeh\} & A cyclist cannot be an emergency vehicle \\
\{not Mobike, not LarVeh\} & A motorbike cannot be a large vehicle\\
\{not Cyc, not TL\} & A cyclist cannot be a traffic light \\
\{not Cyc, not OthTL\} & A cyclist cannot be a other traffic light \\ 
\{not Mobike, not Bus\} & A motorbike cannot be a bus\\
\{not Cyc, not Red\} &  A cyclist cannot signal red \\
\{not MedVeh, not LarVeh\} & A medium vehicle cannot be a large vehicle\\
\{not Mobike, not EmVeh\} & A motorbike cannot be an emergency vehicle \\
\{not Cyc, not Amber\} & A cyclist cannot signal amber\\
\{not Car, not Wait2X\} & A car cannot wait to cross\\
\{not TL, not TurLft\} & A traffic light cannot turn left \\
\{not Green, not MovTow\} & A green traffic light cannot move towards \\ 
\{not Red, not Stop\} & A red traffic light cannot stop \\
\{not OthTL, not IncatRht\} & A other traffic light cannot indicate right \\
\{not TurRht, not TL\} & A traffic light cannot turn right \\
\{not Amber, not Brake\} & A amber traffic light cannot brake \\
\{not OthTL, not HazLit\} & A other traffic light cannot have the hazards light on \\ 
\{not Red, not IncatLeft\} & A red traffic light cannot indicate left \\
\{not Green, not Mov\} & A green traffic light cannot move \\
\bottomrule
    \end{tabular}
    \caption{Requirements table. \label{tab:req_list1}}
    \label{tab:my_label}
\end{table}

\end{landscape}

\begin{landscape}
\begin{table}[]
\centering
\scriptsize
\begin{tabular}{ll}
\toprule
\textbf{Requirements} & \textbf{Natural Language Explanations}\\
\midrule
\{not MedVeh, not Bus\}                                          & A medium vehicle cannot be a bus  \\
\{not Car, not Xing\}                                            & A car cannot be crossing \\
\{not Mobike, not OthTL\}                                        & A motorbike cannot be an other traffic light  \\
\{not Car, not PushObj\}                                         & A car cannot push objects \\
\{not MedVeh, not EmVeh\}                                        & A medium vehicle cannot be an emergency one  \\
\{not Mobike, not Red\}                                          & A motorbike cannot be a red traffic light \\
\{not LarVeh, not Bus\}                                          & A large vehicle cannot be a bus \\
\{not Brake, not Cyc\}                                           & A cyclist cannot brake  \\
\{not MedVeh, not TL\}                                           & A traffic light cannot be a medium vehicle \\
\{not Mobike, not Amber\}                                        & A motorbike cannot be an amber traffic light  \\
\{not LarVeh, not EmVeh\}                                        & A large vehicle cannot be an emergency vehicle \\
\{not Mobike, not Green\}                                        & A motorbike cannot be a green traffic light \\
\{not MedVeh, not OthTL\}                                        & A medium vehicle cannot be an other traffic light \\
\{not HazLit, not Cyc\}                                          & A cyclist cannot have the hazard lights on \\
\{not MedVeh, not Red\}                                          & A medium vehicle cannot be a red traffic light \\
\{not LarVeh, not TL\}                                           & A large vehicle cannot be a traffic light \\
\{not Car, not Ped\}                                             & A car cannot be a pedestrian \\
\{not Mobike, not TL\} & A motorbike cannot be a traffic light\\
\{not Bus, not EmVeh\}                                           & A bus cannot be an emergency vehicle \\
\{not MedVeh, not Amber\}                                        & A medium vechile cannot be an amber traffic light \\
\{not LarVeh, not OthTL\}                                        & A large vehicle cannot be a other traffic light \\
\{not MedVeh, not Green\}                                        & A medium vehicle cannot be a green traffic light \\
\{not Bus, not TL\}                                              & A bus cannot be a traffic light \\
\{not LarVeh, not Red\}                                          & A large vehicle cannot be a red traffic light \\
\{not Bus, not OthTL\}                                           & A bus cannot be a other traffic light \\
\{not LarVeh, not Amber\}                                        & A large vehicle cannot be a amber traffic light \\
\{not Cyc, not PushObj\}                                         & A cyclist cannot push an object \\
\{not EmVeh, not TL\}                                            & An emergency vehicle cannot be a traffic light \\
\{not LarVeh, not Green\}                                        & A large vehicle cannot be a green traffic light \\
\{not Bus, not Red\}                                             & A bus cannot be a red traffic light \\
\{not EmVeh, not OthTL\}                                         & An emergency vehicle cannot be a other traffic light \\
\{not Bus, not Amber\}                                           & A bus cannot be an amber traffic light \\
\{not EmVeh, not Red\}                                           & A emergency vehicle cannot be a red traffic light \\
\{not Mobike, not Wait2X\}                                       & A motorbike cannot wait to cross \\
\{not Bus, not Green\}                                           & A bus cannot be a green traffic light \\
\{not TL, not OthTL\}                                            & A traffic light cannot be a other traffic light \\
\{not EmVeh, not Amber\}                                         & A emergency vehicle cannot be a amber traffic light \\
\{not Mobike, not Xing\}                                         & A motorbike cannot be crossing \\
\{not Cyc, not Ped\}                                             & A cyclist cannot be a pedestrian \\
\{Ped, Cyc, not Xing\}                                           & If an agent is crossing it is either a pedestrian or a cyclist \\
\{not Mobike, not PushObj\}                                      & A motorbike cannot push objects \\
\{not EmVeh, not Green\}                                         & An emergency vehicle cannot be a green traffic light \\
\{not MedVeh, not Wait2X\}                                       & A medium vehicle cannot wait to cross \\
\{not TL, not MovAway\}                                          & A traffic light cannot move away \\
\{not MedVeh, not Xing\}                                         & A medium vehicle cannot be crossing \\
\{not Red, not Amber\}                                           & A red traffic light cannot be amber \\
\{not MedVeh, not PushObj\}                                      & A medium vehicle cannot push objects \\
\{not MovTow, not TL\}                                           & A traffic light cannot move towards \\
\{not OthTL, not MovAway\}                                       & A other traffic light cannot move away \\
\{not LarVeh, not Wait2X\}                                       & A large vehicle cannot wait to cross \\
\{not TL, not Mov\}                                              & A traffic light cannot move  \\
\{not Red, not Green\}                                           & A red traffic light cannot be green \\
\bottomrule
\end{tabular}
\hfill
\begin{tabular}{l l}
\toprule
\textbf{Requirements} & \textbf{Natural Language Explanations}\\
\midrule
\{not TL, not XingFmRht\}                  & A traffic light cannot be crossing from right \\
\{not Amber, not IncatRht\}                & An agent cannot indicate right and signal amber  \\
\{not Red, not TurLft\}                    &  An agent cannot turn left and signal red \\
\{not MovTow, not Mov\}                    & An agent cannot move towards and move  \\
\{not TL, not Xing\}                       & A traffic light cannot cross \\
\{not OthTL, not Wait2X\}                  & A other traffic light cannot wait to cross  \\
\{not Green, not IncatLeft\}               & An agent cannot indicate left and signal green \\
\{not Red, not TurRht\}                    & An agent cannot turn right and signal red \\
\{not Amber, not HazLit\}                  & An agent cannot have the hazard lights on and signal amber \\
\{not TL, not PushObj\}                    & A traffic light cannot push objects \\
\{not OthTL, not XingFmLft\}               & A other traffic light cannot cross from left  \\
\{not Green, not IncatRht\}                & An agent cannot signal green and indicate right  \\
\{not Red, not Ovtak\}                     & An agent cannot signal red and overtake \\
\{not Amber, not TurLft\}                  &  An agent cannot signal amber and turn left \\
\{not OthTL, not XingFmRht\}               & A other traffic light cannot cross from right \\
\{not Red, not Wait2X\}                    & An agent cannot signal red and wait to cross  \\
\{not Green, not HazLit\}                  & An agent cannot signal green and have the hazard lights on  \\
\{not Amber, not TurRht\}                  & An agent cannot signal amber and turn right \\
\{not OthTL, not Xing\}                    & A other traffic light cannot cross \\
\{not Bus, not Ped\}                       & A bus cannot be a pedestrian  \\
\{not Red, not XingFmLft\}                 & An agent cannot signal red and cross from left  \\
\{not OthTL, not PushObj\}                 & A other traffic light cannot push an object  \\
\{not Green, not TurLft\}                  &  An agent cannot signal green and turn left \\
\{not Amber, not Ovtak\}                   &  An agent cannot signal amber and overtake \\
\{not Mov, not Stop\}                      & An agent cannot move and stop  \\
\{not Red, not XingFmRht\}                 & An agent cannot signal red and cross from right   \\
\{not Amber, not Wait2X\}                  & An agent cannot signal amber and wait to cross \\
\{not Green, not TurRht\}                  & An agent cannot signal green and turn right \\
\{not Red, not Xing\}                      & An agent cannot signal and cross  \\
\{not Amber, not XingFmLft\}               & An agent cannot signal amber and cross from left  \\
\{not Green, not Ovtak\}                   & An agent cannot signal green and overtake \\
\{not Amber, not XingFmRht\}               & An agent cannot signal amber and cross from right \\
\{not EmVeh, not Ped\}                     & An emergency vehicle cannot be a pedestrian  \\
\{not Green, not Wait2X\}                  & An agent cannot signal green and wait to cross \\
\{not Amber, not Xing\}                    & An agent cannot signal amber and cross \\
\{not Green, not XingFmLft\}               &  An agent cannot signal green and cross from left \\
\{not Green, not XingFmRht\}               & An agent cannot signal green and cross from right  \\
\{not Green, not Xing\}                    & An agent cannot signal green and cross  \\
\{not TL, not Ped\}                        & A traffic light cannot be a pedestrian  \\
\{not IncatLeft, not IncatRht\}            & An agent cannot indicate left and right \\
\{not OthTL, not Ped\}                     & A other traffic light cannot be a pedestrian \\
\{not Brake, not Wait2X\}                  & An agent cannot brake and wait to cross \\
\{not Red, not Ped\}                       & A pedestrian cannot signal red \\
\{not Xing, not Brake\}                    & An agent cannot cross and brake  \\
\{not Wait2X, not IncatLeft\}              & An agent cannot wait to cross and indicate left \\
\{not Brake, not PushObj\}                 & An agent cannot brake and push an object \\
\{not Amber, not Ped\}                     & A pedestrian cannot signal amber \\
\{not Wait2X, not IncatRht\}               &  An agent cannot wait to cross and indicate right \\
\{not Green, not Ped\}                     &  A pedestrian cannot signal green \\
\{not Wait2X, not Ovtak\}                  &  An agent cannot wait to cross and overtake \\
\{not Wait2X, not XingFmLft\}              & An agent cannot wait to cross and cross from left \\
\{not Wait2X, not XingFmRht\}   & An agent cannot wait to cross and cross from right  \\
\bottomrule
\end{tabular}
\caption{Requirements tables. \label{tab:req_list2}}
\end{table}

\end{landscape}

\begin{landscape}
\begin{table}[]
\centering
\scriptsize
\begin{tabular}{ll}
 \toprule
\textbf{Requirements} & \textbf{Natural Language Explanations}\\
\midrule
\{not Wait2X, not Xing\}                                                                                                      & An agent cannot wait to cross and cross \\
\{not Brake, not Ped\}                                                                                                        &  A pedestrian cannot brake \\
\{not Ped, not IncatLeft\}                                                                                                    & A pedestrian cannot indicate left  \\
\{not Ped, not IncatRht\}                                                                                                     & A pedestrian cannot indicate right  \\
\{not HazLit, not Ped\}                                                                                                       &  A pedestrian cannot have the hazard lights on \\
\{not Cyc, not Green\} & A cyclist cannot signal green \\
\{not OutgoLane, not OutgoCycLane\}                                                                                           & An outgoing lane cannot be an outgoing cycle lane  \\
\{not Ped, not Ovtak\}                                                                                                        & A pedestrian cannot overtake \\
\{not VehLane, not IncomCycLane\}                                                                                             & The vehicle lane cannot be an incoming cycle lane  \\
\{not OutgoLane, not IncomCycLane\}                                                                                           & An incoming cycle lane cannot be a outgoing lane \\
\{not IncomLane, not OutgoCycLane\}                                                                                           & An outgoing cycle lane cannot be an incoming lane \\
\{not OutgoLane, not Pav\}                                                                                                    & An outgoing lane cannot be a pavement \\
\{not IncomCycLane, not OutgoCycLane\}                                                                                        &  An incoming cycle lane cannot be an outgoing cycle lane\\
\{not OutgoLane, not RhtPav\}                                                                                                 & An outgoing lane cannot be a right pavement  \\
\{not IncomCycLane, not Pav\}                                                                                                 & An incoming cycle lane cannot be a pavement \\
\{not XingLoc, not OutgoLane\}                                                                                                &  A crossing cannot be an outgoing lane \\
\{not VehLane, not Parking\}                                                                                                  &  A vehicle lane cannot be a parking \\
\{not BusStop, not OutgoLane\}                                                                                                & A bus stop cannot be a outgoing lane \\
\{not OutgoLane, not Parking\}                                                                                                & An outgoing lane cannot be a parking \\
\{not BusStop, not OutgoCycLane\}                                                                                             &  A bus stop cannot be a outgoing cycle lane \\
\{not Parking, not OutgoCycLane\}                                                                                             & A parking cannot be a outgoing cycle lane \\
\{not XingLoc, not IncomCycLane\}                                                                                             &  A crossing cannot be an incoming cycle lane\\
\{not LftPav, not RhtPav\}                                                                                                    & A pavement is either on the left or on the right  \\
\{not IncomLane, not Parking\}                                                                                                & An incoming lane cannot be a parking \\
\{not IncomCycLane, not BusStop\}                                                                                             &  An incoming cycle lane cannot be a bus stop\\
\{not IncomCycLane, not Parking\}                                                                                             & An incoming cycle lane cannot be a parking \\
\{not Pav, not Parking\}                                                                                                      &  A pavement cannot be a parking \\
\{not LftPav, not Parking\}                                                                                                   & A left pavement cannot be a parking  \\
\{not RhtPav, not Parking\}                                                                                                   & A right pavements cannot be a parking \\
\{not Jun, not Parking\}                                                                                                      & A junction cannot be a parking \\
\{not XingLoc, not BusStop\}                                                                                                  & A crossing cannot be a bus stop \\
\{not XingLoc, not Parking\}                                                                                                  & A crossing cannot be a parking \\
\{not BusStop, not Parking\}                                                                                                  &  A bus stop cannot be a parking\\
\{not Mobike, not Ped \} & A motorbike cannot be a pedestrian \\
\{not MovTow, not OthTL\} & A other traffic light cannot move towards \\
\{not TL, not Brake\} & A traffic light cannot brake \\
\{not Red, not MovAway\} & A red traffic light cannot move away \\ 
\{not Amber, not Green\} & An amber traffic light cannot be green \\
\{not LarVeh, not Xing\} & A large vehicle cannot cross \\
\{not OthTL, not Mov\} & A other traffic light cannot move \\ 
\{not Stop, not TL\} & A traffic light cannot stop \\
\{not LarVeh, not PushObj\} & A large vehicle cannot push objects \\
\{not Red, not MovTow\} & A red traffic light cannot move towards \\
\{not Amber, not MovAway\} & A amber traffic light cannot move away \\
\{not Bus, not Wait2X\} & A bus cannot wait to cross \\
\{not TL, not IncatLeft\} & A traffic light cannot idicate left \\
\{not OthTL, not Brake\} & A other traffic light cannot brake \\
\{not Red, not Mov\} & A red traffic light cannot move \\
\{not TL, not IncatRht\} & A traffic light cannot indicate right \\ 
\{not Stop, not OthTL\} & A other traffic light cannot stop \\ 
\{not Amber, not MovTow\} & A amber traffic light cannot move towards \\ 
\{not Green, not MovAway\} & A green traffic light cannot move away \\
\{not TL, not HazLit\} & A traffic light cannot have the hazard lights on \\
\{not Red, not Brake\} & A red traffic light cannot brake \\
\{not Bus, not Xing\} & A bus cannot be crossing \\ 
\{not OthTL, not IncatLeft\} & A other traffic light cannot indicate right \\
\{not MedVeh, not Ped\} & A medium vehicle cannot be a pedestrian \\
\{not Amber, not Mov\} & A amber traffic light cannot move \\
\{not Bus, not PushObj\} & A bus cannot push objects \\ 
\{not EmVeh, not Wait2X\} & A emergency vehicle cannot wait to cross \\
\bottomrule
\end{tabular}
\hfill
\begin{tabular}{ll}
 \toprule
\textbf{Requirements} & \textbf{Natural Language Explanations}\\
\midrule
\{not TL, not Ovtak\} & A traffic light cannot overtake \\
\{not Amber, not Stop\} & A amber traffic light cannot stop\\
\{not EmVeh, not Xing\} & An emergency vehicle cannot be crossing \\
\{not OthTL, not TurLft\} & A other traffic light cannot turn left \\
\{not Red, not IncatRht\} & A red traffic light cannot indicate right  \\
\{not TL, not Wait2X\} & A traffic light cannot wait to cross \\
\{not Green, not Brake\} & A green traffic light  cannot be green and break \\
\{not MovAway, not Mov\} & An agent cannot move perpendicularly to and away from the AV \\
\{not EmVeh, not PushObj\} & An emergency vehicle cannot push an object \\
\{not TurRht, not OthTL\} & A other traffic light cannot turn right \\
\{not Amber, not IncatLeft\} & An amber traffic light cannot indicate left\\
\{not TL, not XingFmLft\} & An traffic light cannot be crossing from the left\\
\{not Red, not HazLit\} & A red traffic light cannot have the hazard lights on \\
\{not Green, not Stop\} & A green traffic light  cannot stop \\
\{not LarVeh, not Ped\} & A large vehicle cannot be a pedestrian \\
\{not OthTL, not Ovtak\} & A other traffic light cannot overtake \\
\bottomrule
\\\\\\\\\\\\\\\\\\\\\\\\\\\\\\\\\\\\\\\\\\\\\\\\\\\\\\\\\\\\\\\\\\\\\\\\\\\\\\\\\\\\\\\\
\end{tabular}
\caption{Requirements tables. \label{tab:req_list3}}
\end{table}
\end{landscape}

\end{document}